\documentclass[manuscript,screen]{acmart}
\usepackage{multirow}
\usepackage{array}
\usepackage{pgfplots}
\usepackage{tikz}
\usepackage{subcaption}
\pgfplotsset{compat=1.18}
\usepgfplotslibrary{fillbetween}

\definecolor{colorAcc}{RGB}{54, 95, 145}
\definecolor{colorF1}{RGB}{146, 169, 189}

\newcommand{\subfloat}[2][]{\subcaptionbox{}{\centering #2}}

\AtBeginDocument{%
  }

\setcopyright{none}
\copyrightyear{2026}
\acmYear{2026}
\acmDOI{}
\acmJournal{TALLIP}

\begin{document}

\title{Rationale-Guided Knowledge Distillation for Cross-Lingual Stance Detection}
\thanks{Research results of Anhui Province Philosophy and Social Science Planning Project (No. AHSKQ2024D044).}

\author{Qiuli Zhou}
\email{zhouql@hfut.edu.cn}
\affiliation{%
  \institution{School of Foreign Studies, Hefei University of Technology}
  \city{Hefei}
  \postcode{230601}
  \country{China}
}

\author{Jingyuan Yao}
\email{2024212237@mail.hfut.edu.cn}
\affiliation{%
  \institution{School of Computer Science and Information Engineering, Hefei University of Technology}
  \city{Hefei}
  \postcode{230601}
  \country{China}
}

\author{Shengeng Tang}
\authornote{Corresponding author.}
\email{tangsg@hfut.edu.cn}
\affiliation{%
  \institution{School of Computer Science and Information Engineering, Hefei University of Technology}
  \city{Hefei}
  \postcode{230601}
  \country{China}
}

\author{Hongzhi Chen}
\email{chenhz313@126.com}
\affiliation{%
  \institution{School of Artificial Intelligence Engineering, Hefei Institute of Technology}
  \city{Hefei}
  \postcode{238076}
  \country{China}
}

% \author{Weidong Chen}
% \email{chenweidong@ustc.edu.cn}
% \affiliation{%
%   \institution{School of Information Science and Technology, University of Science and Technology of China}
%   \city{Hefei}
%   \postcode{230000}
%   \country{China}
% }

\author{Jun Tang}
\authornotemark[1]
\email{958748910@qq.com}
\affiliation{%
  \institution{School of Foreign Studies, Hefei University of Technology}
  \city{Hefei}
  \postcode{230601}
  \country{China}
}

\author{Richang Hong}
\email{hongrc@hfut.edu.cn}
\affiliation{%
  \institution{School of Computer Science and Information Engineering, Hefei University of Technology}
  \city{Hefei}
  \postcode{230601}
  \country{China}
}

\renewcommand{\shortauthors}{Zhou et al.}

% \begin{abstract}
% Stance detection, the task of determining a user's attitude (e.g. favor or against) within a text toward a specific target, is of significant value in public opinion mining, market analysis and many other fields. While extensive research has been conducted in monolingual settings, particularly for English, many languages such as Catalan lack sufficient resources to develop high-performance models. Cross-lingual stance detection addresses this challenge by transferring knowledge from high-resource to low-resource languages. However, existing methods in this area primarily prioritize superficial semantic correlation between texts and targets, overlooking the underlying logical reasoning indispensable for robust stance inference. Although certain methods incorporate LLM to bridge this gap, their high resource demands limit their feasibility in realistic settings. In our paper, we propose a knowledge distillation framework which leverages the reasoning capabilities of Large Language Models (LLM) to train a smaller, more efficient student model for improved stance detection.   Additionally, we incorporate two different contrastive learning methods to improve the model's discrimination ability between opposing stances. Experiments on multilingual datasets demonstrate the effectiveness of our method compared with competitive baselines.
% \end{abstract}

\begin{abstract}
Stance detection aims to identify whether a text expresses a favorable or opposing attitude toward a given target, and serves as an important task for various downstream applications. Although existing studies have achieved strong performance in monolingual settings, especially in English, many low-resource languages such as Catalan still lack sufficient annotated data for training effective models. Cross-lingual stance detection alleviates this problem by transferring stance knowledge from resource-rich languages to low-resource languages. However, most existing methods mainly rely on semantic alignment between texts and targets, while ignoring the reasoning process required for reliable stance inference. Although Large Language Models provide strong reasoning ability, their high computational cost and inference latency limit practical deployment. To address these limitations, we propose a rationale-guided knowledge distillation framework for cross-lingual stance detection. Specifically, we use Chain-of-Thought prompting to guide Large Language Models in generating informative rationales, and distill the resulting reasoning knowledge into a compact student model. We further design a dual-path distillation mechanism to align rationale-enhanced and rationale-free representations, together with their prediction distributions. In addition, two contrastive learning strategies are introduced to improve stance discrimination. Experiments on multilingual benchmarks demonstrate that our method consistently outperforms competitive baselines.
\end{abstract}




\maketitle

\section{Introduction}

% With the rapid proliferation of social media platforms, stance detection, which aims to identify a user's attitude (e.g., favor, neutral, or against) within a text toward a specific target (e.g., topic or claim) has demonstrated immense research potential \cite{socialmedia}. In recommendation systems, this task serves as a critical mechanism for analyzing user preferences and managing content diversity. By assessing the alignment or opposition between users and content, recommendation engines can better balance the diversity of viewpoints presented to users and mitigate the risk of reinforcing selective exposure \cite{recom1,recom2}.

With the rapid growth of social media platforms, stance detection has attracted increasing attention in natural language processing. This task aims to identify a user's attitude, such as favor, neutral, or against, toward a specific target, such as a topic or claim, from a given text \cite{socialmedia}. It supports a wide range of downstream applications. For example, in recommendation systems, stance detection can help analyze user preferences and manage content diversity. By identifying whether users support or oppose specific content, recommendation models can better balance viewpoint diversity and reduce the risk of reinforcing selective exposure \cite{recom1,recom2}.

% Existing work has predominantly focused on monolingual settings, with a heavy reliance on English datasets \cite{mono1,sdsurvey,vast}. In contrast, datasets for other languages are often limited in size and scope of topics, which diminishes the accuracy and generalization ability of stance detection models in these languages \cite{mono3}. To address these data imbalances, cross-lingual stance detection has emerged as a strategy to transfer knowledge from high-resource to low-resource settings, finally transferring the detection capabilities from high-resource languages to low-resource settings \cite{zhang2023cross,zhang2023target,zhang2024llm}. For instance, \cite{zhang2023cross} proposed a cross-lingual and cross-target distillation framework (CCSD) that mines the category information through target representation to bridge the target inconsistency gap while also constructing cross-lingual stance templates to achieve cross-lingual knowledge transfer. While \cite{zhang2024llm} proposed leveraging external background knowledge and inference knowledge from high-resource languages, this approach relies on external sources that are difficult to generalize to unseen targets in target languages, as constructing sufficiently comprehensive knowledge bases for diverse topics remains a significant challenge.

Most existing studies focus on monolingual settings, especially English \cite{mono1,sdsurvey,vast}. In contrast, many other languages have limited annotated data and cover only a narrow range of topics, which restricts the performance and transfer ability of stance detection models \cite{mono3}. To address this data imbalance, cross-lingual stance detection transfers stance knowledge from resource-rich languages to low-resource languages \cite{zhang2023cross,zhang2023target,zhang2024llm}. For example, \cite{zhang2023cross} proposed a cross-lingual and cross-target distillation framework, namely CCSD, which mines category information from target representations to reduce target inconsistency and constructs cross-lingual stance templates for knowledge transfer. In addition, \cite{zhang2024llm} introduced external background knowledge and inference knowledge from resource-rich languages. However, this strategy depends on external resources that may be difficult to generalize to unseen targets, because constructing comprehensive knowledge bases for diverse topics remains challenging.

\begin{figure*}[tbp]
    \centering
    \includegraphics[width=\linewidth]{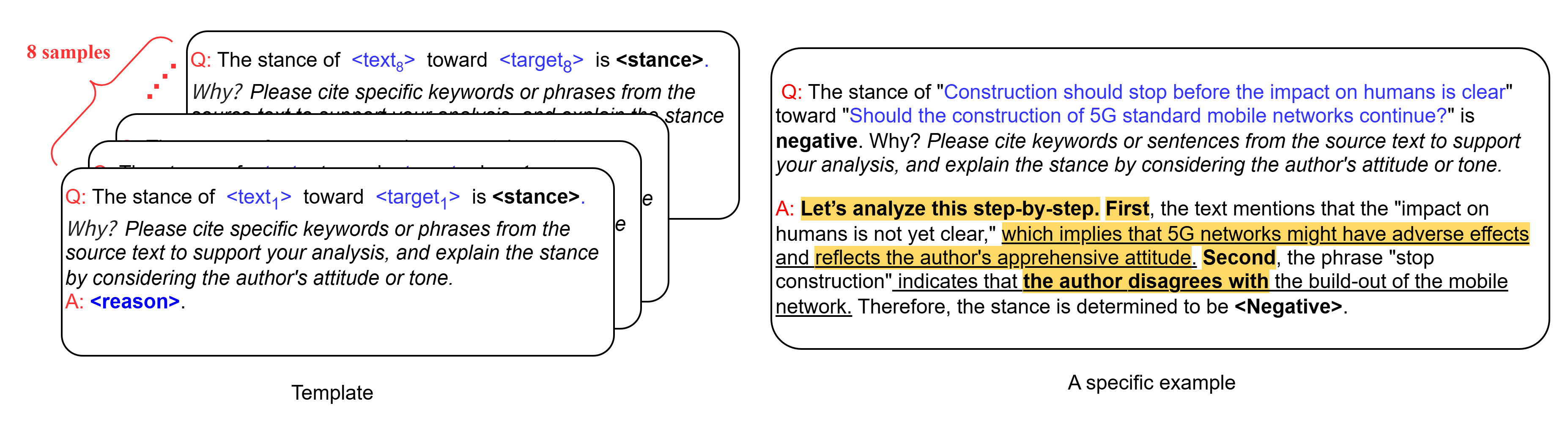}
    \caption{Illustrative example of Chain-of-Thought (CoT) reasoning for stance detection in English.}
    \Description{Example prompt and Chain-of-Thought rationale used for stance detection.}
    \label{fig:prompt}
\end{figure*}

% Recently, the scaling of language models has led to the emergence of advanced reasoning capabilities,  providing a brand new perspective for cross-lingual knowledge transfer and the improvement of detection accuracy \cite{cotreason,teachreason,emergent,zhou2025hierarchical}. Unlike previous deep learning methods, which primarily aimed to enhance performance by designing complex network architectures \cite{zhang2023target}, human reasoning typically involves identifying the underlying emotions conveyed beneath the surface text. Accordingly, we leverage the reasoning proficiency of LLMs to extract the key components essential for sentiment detection and analyze the author's emotional stance with the nuance of a human expert. As demonstrated in previous studies, these reasoning abilities can be unlocked through chain-of-thought (CoT) prompting, which encourages language models to break down a complex task into a series of intermediate steps \cite{cotreason,teachreason,distillstepbystep,symbolic}. In this work, we incorporate CoT into our framework and adopt in-context learning \cite{incontextsurvey,rethinkingincontext} to generate a logically rigorous reasoning process that the LLM can emulate, as illustrated in Figure~\ref{fig:prompt}. This approach leverages instructions to elicit more consistent logic from the model. To facilitate effective learning from these examples, we provide eight demonstrations which cover different topics, ranging from health to political policy, to ensure the generalization ability of our model.

Recent advances in Large Language Models have shown strong reasoning capabilities, providing a new direction for cross-lingual knowledge transfer and stance inference \cite{cotreason,teachreason,emergent,zhou2025hierarchical}. Different from previous neural methods that mainly improve performance through model architecture design \cite{zhang2023target}, human stance judgment often requires identifying implicit emotions, attitudes, and reasoning cues behind the surface text. Motivated by this observation, we use the Large Language Model (LLM) to extract stance related evidence and generate rationales that explain the author's attitude toward the target. Prior studies have shown that such reasoning abilities can be elicited through Chain-of-Thought prompting, which guides language models to decompose a complex task into intermediate reasoning steps \cite{cotreason,teachreason,distillstepbystep,symbolic}. In this work, we combine Chain-of-Thought prompting with In-Context Learning \cite{incontextsurvey,rethinkingincontext} to generate coherent rationales for stance detection, as shown in Fig~\ref{fig:prompt}. Specifically, we provide 8 demonstrations covering diverse topics, such as health and political policy, to encourage the model to produce consistent and transferable reasoning patterns.

Although Large Language Models are effective in reasoning, directly deploying them for stance detection is often impractical due to high computational cost and inference latency. Their large model size makes it difficult to satisfy real-time response requirements in practical applications \cite{meituan}. To improve efficiency, we adopt knowledge distillation \cite{hinton2015distilling,kdsurvey} to transfer the reasoning ability of Large Language Models to a compact model. Since the BERT family \cite{bert,mbert} has achieved strong performance on many downstream tasks, and its encoder architecture is suitable for cloze-style stance prediction, we use multilingual BERT (mBERT) as the student model. Specifically, we fine-tune mBERT with rationales generated by the LLM, enabling the student model to learn reasoning-aware representations for stance detection.

However, directly distilling knowledge from Large Language Models to mBERT is nontrivial. Large Language Models are usually based on decoder-only or encoder-decoder architectures, whereas mBERT follows an encoder only structure. This architecture gap makes direct output alignment difficult \cite{mismatch}. To address this problem, we design a dual-branch hierarchical distillation framework, which contains a rationale-enhanced path and a standard path. The rationale-enhanced path takes the target, text, and LLM-generated rationale as input, while the standard path only takes the target and text as input. By aligning both hidden representations and prediction distributions between the two paths, the student model can learn reasoning knowledge during training while maintaining efficient rationale-free inference at test time.

\begin{figure*}[tbp]
    \centering
    \includegraphics[width=\linewidth]{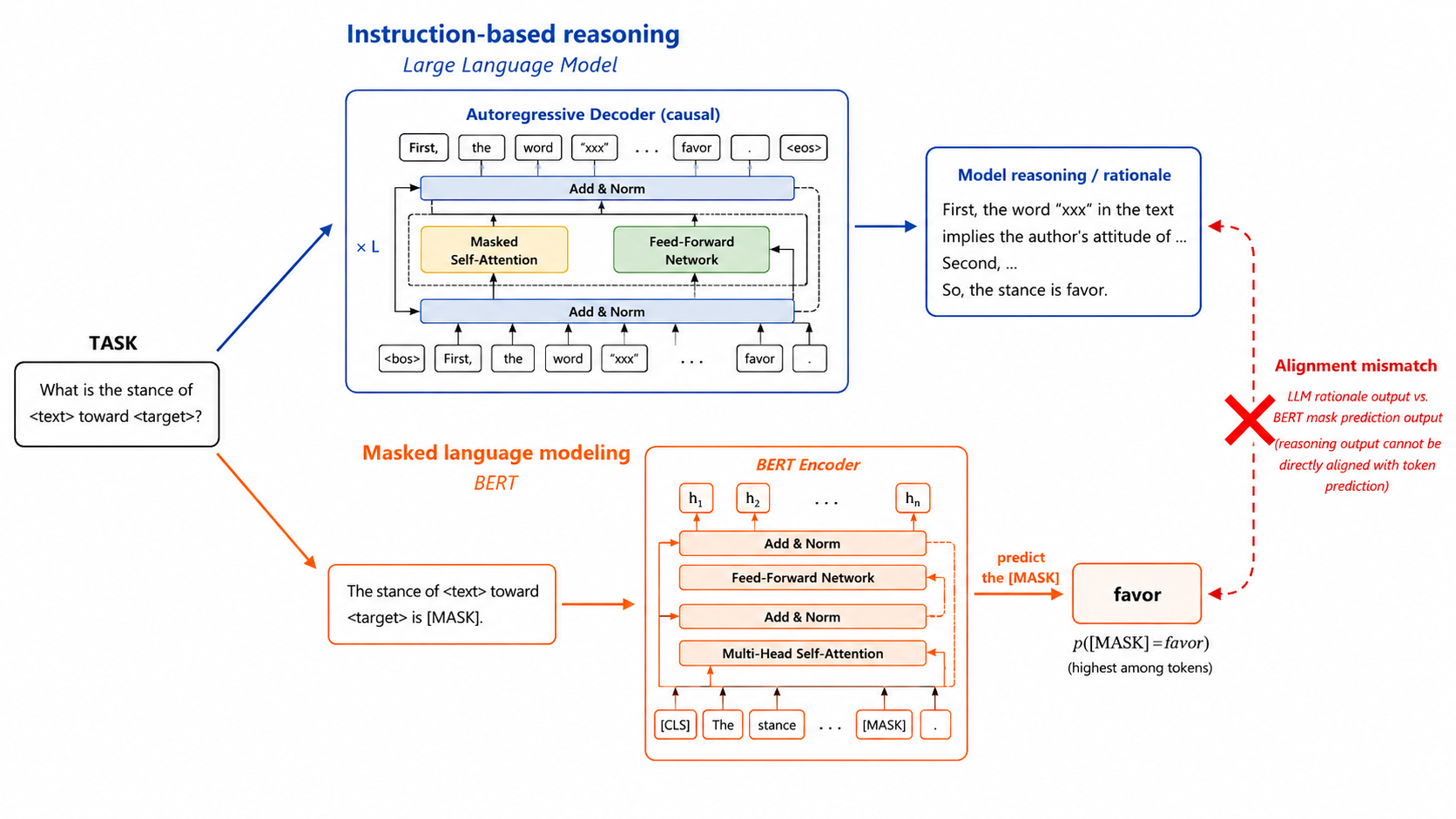}
    \caption{Representation and structural gaps in knowledge distillation between decoder-only Large Language Models and encoder-only mBERT.}
    \Description{Diagram showing representation and architecture gaps between decoder-only language models and encoder-only mBERT.}
    \label{fig:mismatch}
\end{figure*}

% To further enhance the distinguishing ability of mBERT between opposing stances, we introduce two different contrastive learning methods \cite{gao2021simcse,con3,han2023text,tian2019contrastive} to our Source-Language Training step and Teacher-Student Distillation step. In source language settings, the data are sufficient, so we construct class prototypes based on support samples and calculate the distance of query samples to prototypes as contrastive loss. For the target language, due to the low-resource problem, where the training data is about 10 times less than the source language, we do not construct class prototypes, instead we design a dual-constraint contrastive alignment which aligns the embeddings of the same sample in two routes and pushes those of different samples from different routes. As a result, the student model can strengthen its ability to discriminate between opposing stances.

To further improve the discrimination between opposing stances, we introduce two contrastive learning strategies in the source-language training stage and the teacher-student distillation stage \cite{gao2021simcse,con3,han2023text,tian2019contrastive}. In the source-language stage, where sufficient labeled data are available, we construct class prototypes from support samples and optimize query representations according to their distances to the stance prototypes. In the target-language stage, where training data are limited, we avoid constructing unstable prototypes. Instead, we design a dual constraint contrastive alignment objective that pulls together the two path representations of the same sample and pushes apart representations from different samples. This design helps the student model capture stance discriminative features under low-resource conditions.
We conduct extensive experiments on multilingual datasets covering diverse topics. The results show that our method consistently outperforms competitive baselines, demonstrating the effectiveness of rationale-guided knowledge distillation for cross-lingual stance detection.

% We conduct extensive experiments on multilingual datasets covering a wide range of topics, and the results demonstrate the effectiveness of our method compared with competitive baselines.

The main contributions of our work are as follows:
\begin{itemize}

% \item \textbf{A Novel Cross-Lingual Reasoning Framework.}  We propose a framework that leverages the emergent reasoning capabilities of Large Language Models (LLMs) via Chain-of-Thought (CoT) prompting. By utilizing instructions to guide the analysis of target-language content, our approach effectively elicits rigorous logical rationales to enhance stance detection in low-resource settings.

\item \textbf{A rationale-guided cross-lingual reasoning framework.} We propose a framework that leverages the reasoning ability of Large Language Models through Chain-of-Thought prompting. By using instruction-based rationale generation, the proposed method provides explicit reasoning supervision for low-resource cross-lingual stance detection.

% \item \textbf{Dual-Branch Hierarchical Distillation for Architectural Compatibility.}  To overcome the structural mismatch between decoder-based LLMs and encoder-based student models (mBERT) and achieve more efficient knowledge transfer, we incorporate a dual-branch hierarchical distillation mechanism. This architecture allows the student model to acquire rich inferential knowledge from rationales during training while maintaining high efficiency by simulating a rationale-free environment for inference through both the representation-level and the response-level distillation.

\item \textbf{Dual-branch hierarchical distillation for architecture compatibility.} To bridge the structural gap between decoder-based Large Language Models and encoder-based mBERT, we design a dual-branch distillation mechanism. This framework enables the student model to learn from rationale-enhanced inputs during training while supporting efficient rationale-free inference through representation-level and response-level alignment.

% \item \textbf{Enhanced Knowledge Transfer via Heterogeneous Contrastive Constraints.}  We incorporate different contrastive learning methods in the training stages of source language and target language to supplement traditional label alignment. This optimization enables the student model to better discriminate between opposing stances and more effectively internalize the latent features of the reasoning process.

\item \textbf{Enhanced stance transfer with contrastive constraints.} We introduce different contrastive learning objectives for the source-language and target-language stages. These constraints complement label supervision and improve the model's ability to distinguish between opposing stances.

\end{itemize}

\section{Related Work}

This section reviews three lines of research that are closely related to our work: stance detection, knowledge distillation, and LLM reasoning. These three areas correspond to the task background, the transfer mechanism, and the source of reasoning knowledge in our framework, respectively. First, stance detection provides the fundamental task setting, where the main research focus has gradually shifted from supervised monolingual classification to cross-target and cross-lingual scenarios under data scarcity. This line of work motivates the need for effective knowledge transfer in low-resource languages. Second, knowledge distillation offers a practical solution for transferring useful knowledge from large or specialized teacher models to compact student models, which is essential for improving efficiency and reducing the deployment cost of stance detection systems. Third, recent studies on LLM reasoning show that Large Language Models can generate intermediate rationales through In-Context Learning and Chain-of-Thought prompting, providing richer logical supervision than conventional label-based learning. Taken together, these studies motivate our rationale-guided knowledge distillation framework: LLM reasoning supplies explicit stance rationales, knowledge distillation transfers such reasoning knowledge into an efficient mBERT-based student model, and cross-lingual stance detection serves as the target application.

\subsection{Stance Detection}

Stance detection initially focused on single-language settings \cite{single1,single2,single3,single4}, with early research primarily established within highly constrained supervised learning environments. In these configurations, models are typically trained and tested on a predefined and fixed set of targets \cite{single2,tan,before2}.

During the supervised learning phase, stance targets were generally confined to noun phrases representing specific entities. This is exemplified by foundational datasets such as SemEval-2016 \cite{single2}, which focuses on predefined entities, and WT-WT \cite{single4}, which centers on financial mergers. A critical limitation of these early datasets is their heavy reliance on specific targets, leading to significant performance degradation when models encounter unseen topics. This bottleneck eventually prompts the research community to pivot toward more flexible cross-target configurations to enhance model generalization \cite{fu2024sentiment,TALLIP_5,chen2021cascade,jin2024improving,tian2023end,ye2025improving,ye2025multi}.

Cross-target stance detection aims to leverage knowledge gained from a source target to identify sentiments regarding a different, yet related, destination target \cite{TALLIP_1}. To achieve this, researchers have developed various strategies to transcend target-specific constraints. For example,  \cite{zhang2023target} utilizes a target relation graph to capture semantic associations between targets both within and across languages. By employing a graph-based approach and relation alignment strategies, the model can effectively bridge the gap between different language domains \cite{wang2025combatting,liu2025matching,zhao2023difference}.

The development of cross-lingual stance detection primarily focuses on overcoming the imbalance of linguistic resources, undergoing a paradigm shift from early machine translation alignment to multilingual pre-trained models and adversarial learning \cite{zhang2023cross,zhang2023target,tian2023end,zhang2024llm,cross1,cross2,TALLIP_2,TALLIP_3,TALLIP_6,song2025towards,chen2026creatiparser,chen2026face}.   In recent years, with the rise of Large Language Models, cross-lingual stance detection has been evolving from simple label alignment toward logical reasoning alignment. Modern frameworks like KEAR \cite{zhang2024llm} have broken the dependence on labeled data in target languages by inducing language-agnostic external knowledge (such as background facts) from the LLM to serve as a bridge, enhancing prediction accuracy in zero-shot settings \cite{wang2023improving,chen2023weakly,jin2024improving,li2025rethinking}.

Despite these advances, most existing stance detection methods still emphasize target-level or language-level alignment through labels, templates, or representations, while the intermediate reasoning process that connects the text, target, and stance label is rarely modeled explicitly. In contrast, our work introduces LLM-generated rationales as transferable reasoning supervision, enabling the student model to learn reasoning-aware stance representations rather than relying only on surface semantic alignment.

\subsection{Knowledge Distillation}
Knowledge distillation focuses on transferring expertise from a large, complex model to a more lightweight counterpart \cite{hinton2015distilling,ye2024dual,kdsurvey,TALLIP_4}. In practical scenarios where computing resources are limited and low latency is required, these techniques are widely used to improve model efficiency, support multilingual and cross-target transfer, and alleviate bias issues in training data. The core idea is to use soft labels or intermediate representations generated by high-performance teacher models to guide the learning process of lightweight or domain-specific student models.

In stance detection, knowledge distillation has been adopted to transfer semantic and task-specific knowledge into more efficient architectures. For instance, BERTtoCNN \cite{kd1} combines traditional distillation loss with similarity-preserving loss to inject BERT's semantic understanding ability into a Text-CNN architecture, significantly improving inference speed on benchmark tasks such as SemEval-2016. Furthermore, CCSD \cite{zhang2023cross} adopts dual-knowledge distillation with a ``Cross-lingual Teacher'' and a ``Cross-target Teacher'', which extract stance category information from source-language data and provide pseudo labels for training the target-language student model \cite{wang2023contour,chen2026subjective,qin2025query,guo2025emoverse}.

Nevertheless, existing distillation methods for stance detection mainly transfer logits, intermediate representations, or pseudo labels from teacher classifiers, which makes it difficult to convey the logical evidence behind stance decisions. Our framework addresses this limitation by distilling rationale-enhanced knowledge: LLM-generated explanations enrich the target-language training inputs, while the dual-path representation-level and response-level alignment enables mBERT to internalize such reasoning knowledge while remaining rationale-free at inference time \cite{ye2024dual}.

\subsection{LLM reasoning}
The evolution of reasoning abilities in Large Language Models  has driven a shift in stance detection from traditional supervised classification toward generative logical reasoning \cite{llm1,llm2,llm3}. The  transition primarily lies in In-Context Learning \cite{incontextsurvey,teachreason,brown2020languageicl,iclexplain,liu2024bootstrapping,chen2022multi,huang2025graph,lin2024prompting}, a mechanism that allows models to perform specific tasks through zero-shot or few-shot prompting according to input instructions without updating parameters. Instruction-tuning techniques further bolster the ability of models to follow complex logical intents, enabling them to capture subtle semantic associations between texts and targets through transfer learning from large-scale corpora \cite{meituan,zhang2025creatidesign,li2024exploring,fu2024sentiment}.

In practical applications, Large Language Models employ Chain-of-Thought \cite{cotreason,cotzero} reasoning to generate intermediate analytical steps, which not only significantly enhances prediction accuracy on complex topic sets like VAST \cite{vast}, but also provides interpretable logical support for decision-making. In cross-lingual settings, the reasoning power of Large Language Models is employed for knowledge mining and alignment; for instance, the RCCA \cite{cross2} framework utilizes reinforcement learning to perform cross-lingual Chain-of-Thought alignment, ensuring logical consistency when processing texts from diverse cultural backgrounds \cite{zhang2026stimuli,wang2026multi}.

However, many LLM reasoning approaches either deploy Large Language Models directly for prediction or require additional reasoning alignment during inference, which can introduce high latency and computational cost in practical low-resource multilingual scenarios. Our method uses the LLM only as an rationale generator, and then transfers the resulting Chain-of-Thought supervision into a compact mBERT student, preserving the benefit of explicit reasoning without requiring LLM inference during deployment.

\section{Methods}

\begin{figure*}[tbp]
    \centering
    \includegraphics[width=\linewidth]{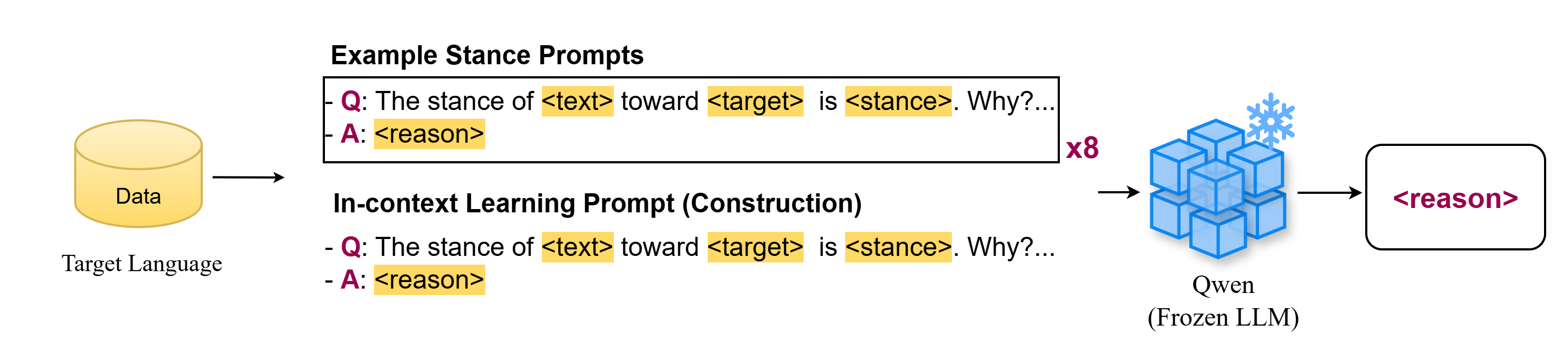}
    \caption{The pipeline of reasoning rationale preparation via In-Context Learning (ICL).}
    \Description{Pipeline for preparing reasoning rationales with In-Context Learning.}
    \label{fig:step0}
\end{figure*}

\begin{figure*}[tbp]
    \centering
    \includegraphics[width=\linewidth]{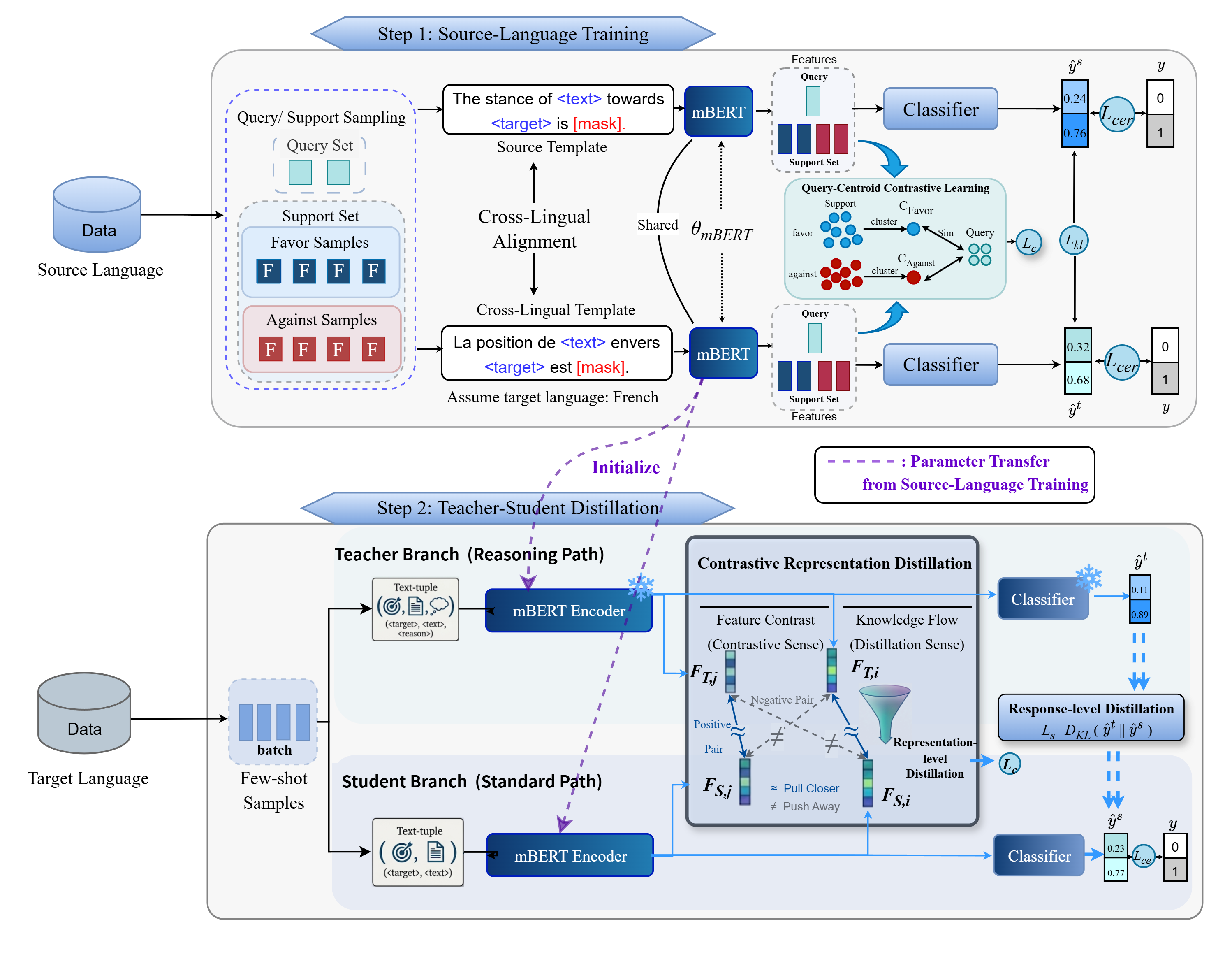}
    \caption{Overall architecture of the proposed framework. Step 1 trains the source-language mBERT model with cross-lingual templates and query-centered contrastive learning. Step 2 trains the target-language student model by dual-path teacher-student distillation, where LLM-generated rationales provide additional supervision and both representation-level and response-level knowledge are transferred.}
    \Description{Two-step architecture for source-language training and target-language teacher-student distillation.}
    \label{fig:step12}
\end{figure*}

Our proposed method first prepares LLM-generated rationales, followed by a training step in the source language and a teacher-student distillation step, as illustrated in Fig~\ref{fig:step0} and Fig~\ref{fig:step12}. During the rationale preparation phase, we employ an In-Context Learning strategy \cite{incontextsurvey,teachreason,brown2020languageicl,iclexplain} to elicit inferential rationales from a frozen Large Language Model (LLM), which are subsequently utilized to guide distillation in the target language. In the source-language training stage, we partition the dataset into a support set and a query set with a 4:1 ratio. The mBERT model is fine-tuned using cross-lingual templates, where class prototypes are constructed based on the output embeddings of the support set for each category. By calculating the distances between query samples and these class prototypes, we derive a contrastive loss to optimize both mBERT and the classifier, thereby enhancing the model's ability to distinguish between opposing stances. In the knowledge distillation stage, to address the challenge of data scarcity, we augment the limited training data with previously obtained LLM rationales to facilitate reasoning-level knowledge transfer. Within this stage, we design a contrastive representation distillation mechanism that pulls together embeddings of the same sample from dual paths while pushing apart those from different samples, thus maximizing the utility of low-resource target-language data. In addition, we design a response-level distillation to enable knowledge transfer of the soft labels.

\subsection{Task Definition}

We define the training dataset in the source language as $D_{\text{src}}=\{t_i,c_i,y_i\}_{i=1}^{N_{src}}$, where $t_i$ denotes the target, $c_i$ represents the input text, $y_i \in \{\text{favor, against}\}$ designates the ground-truth stance label, and $N_{src}$ is the total number of source samples. To facilitate contrastive learning, we further partition the source training data into a support set $D_s=\{t_i,c_i,y_i\}_{i=1}^{N_{\text{sup}}}$ and a query set $D_q=\{t_i,c_i,y_i\}_{i=1}^{N_{\text{q}}}$, maintaining a sample ratio of 4:1 (support samples are divided evenly for each category) between $N_{\text{sup}}$ and $N_q$.

In target languages, however, a significant challenge is the scarcity of high-quality labeled resources.  To mitigate this, we augment the target-language training data by generating a comprehensive stance reasoning process for each training pair $(t_i, c_i)$, thereby providing enriched supervision signals for mBERT fine-tuning. The target language dataset is denoted as $D_{\text{tgt}}=\{t_i,c_i,r_i,y_i\}_{i=1}^{N_t}$, where $r_i$ refers to the inferential rationale generated by the LLM. Notably, the target language scale $N_t$ is significantly smaller than that of the source language, typically by a factor of about 8 to 10. We employ Qwen \cite{qwen3.5} as the base Large Language Model (LLM) to generate these inferential rationales, keeping its parameters frozen throughout the inference phase.

To ensure the quality and consistency of the generated rationales, we adopt an In-Context Learning (ICL) strategy \cite{incontextsurvey,teachreason,brown2020languageicl,iclexplain} to guide the LLM's inferential path. Specifically, we manually construct 8 diverse prompt examples in the target language. The inquiry prompt is structured as follows (translated for clarity): ``\textit{The stance of <text> toward <target> is <stance>. Why? Please cite specific keywords or phrases from the source text to support your analysis, and explain the stance by considering the author's attitude or tone.}'' For the model's responses, we design a Chain-of-Thought (CoT) reasoning paradigm, exemplified by: ``\textit{Let's think step by step. First, the text mentions ..., which implies ... and reflects the author's ... attitude. Second, ...}'' (a specific example is illustrated in Fig~\ref{fig:prompt}). Notably, to enhance the generalization capability of the student model (mBERT), these 8 examples cover a broad spectrum of topics.

% Through this structured learning from varied examples, the LLM is capable of reasoning systematically and identifying the specific linguistic elements most relevant to stance detection. Consequently, the LLM-generated rationales highlight critical semantic features, which subsequently enhance the sensitivity of mBERT to the pivotal information required for cross-lingual stance tasks.

\subsection{Source-Language Training}

In cross-lingual stance detection, it is essential to first fine-tune a model on source-language data in order to transfer relevant knowledge to the target language. We adopt multilingual BERT (mBERT) \cite{bert,mbert} as our backbone due to its strong contextual understanding and its demonstrated effectiveness on downstream tasks such as stance detection. To promote cross-lingual representational capability from the outset, we fine-tune mBERT simultaneously on both source-language and cross-lingual templates. The template structure follows the pattern (illustrated for clarity): \textit{``The stance of <text> towards <target> is [MASK]''}, where \textit{<text>} and \textit{<target>} serve as slot placeholders, and the \textit{[MASK]} token corresponds to the stance label to be predicted. For the target language, the instruction portion of the template is translated accordingly while the prediction mechanism over the masked token remains unchanged.

To further improve the model's ability to discriminate between opposing stances, we partition the source training data into a support set \(D_S\) and a query set \(D_q\). Class prototypes are computed from the support samples of each stance category to characterize category centroids, which are subsequently employed to define a contrastive loss on the query samples.

Specifically, we first extract the last hidden state corresponding to the \texttt{[MASK]} token from the source-language template:
\begin{equation}
    \mathbf{h}_i^s = \text{mBERT}(x_i^s)_{[\text{MASK}]}
\end{equation}
where \(x\) corresponds to the template sequence. Analogously from the cross-lingual template:
\begin{equation}
    \mathbf{h}_i^t = \text{mBERT}(x_i^t)_{[\text{MASK}]}
\end{equation}

Within each batch, we then aggregate the support samples belonging to the same stance category to compute a category prototype, defined as the mean of the corresponding embedding vectors. For the \textit{Favor} category we obtain:
\begin{equation}
    \mathbf{C}_{\text{Favor}} = \frac{1}{|D_{S,\text{favor}}|} \sum_{\mathbf{h}_i \in D_{S,\text{favor}}} \mathbf{h}_i
\end{equation}
and for the \textit{Against} category:
\begin{equation}
    \mathbf{C}_{\text{Against}} = \frac{1}{|D_{S,\text{against}}|} \sum_{\mathbf{h}_i \in D_{S,\text{against}}} \mathbf{h}_i
\end{equation}
where \(\mathbf{h}_i\) collectively denotes both \(\mathbf{h}_i^s\) and \(\mathbf{h}_i^t\), and \(|D_{S,\text{category}}|\) indicates the number of support samples of the given category within the current batch.

Once the category centroids are established, we measure the similarity between each query sample \(\mathbf{h}_q \in D_q\) and the prototypes via cosine similarity:
\begin{equation}
    s(\mathbf{h}_q, \mathbf{C}_k) = \frac{\mathbf{h}_q \cdot \mathbf{C}_k}{\|\mathbf{h}_q\| \, \|\mathbf{C}_k\|}, \quad k \in \{\text{Favor, Against}\}
\end{equation}
where \(s(\mathbf{h}_q, \mathbf{C}_k)\) represents the cosine similarity between the query embedding and the prototype of category \(k\). The contrastive loss for a query sample is then defined as its similarity to the prototype of the \textit{opposite} stance:
\begin{equation}
\mathcal{L}_{\text{c}}(\mathbf{h}_q) =
\begin{cases}
s(\mathbf{h}_q, \mathbf{C}_{\text{Against}}), & \text{if } y_q = \text{Favor} \\
s(\mathbf{h}_q, \mathbf{C}_{\text{Favor}}), & \text{if } y_q = \text{Against}
\end{cases}
\end{equation}
with \(y_q\) denoting the ground-truth label of the query sample.

In parallel with the contrastive module, the embeddings derived from mBERT are passed through a classifier to produce predicted stance probabilities. In our framework, the classifier is implemented as a two-layer feed-forward neural network:
\begin{equation}
    \hat{y}_i = \text{Classifier}(\mathbf{h}_i)
\end{equation}
where \(\hat{y}_i\) refers to the prediction for either the source or cross-lingual template (i.e., \(\hat{y}_i^s\) or \(\hat{y}_i^t\)).

To enforce consistency between the predictions from the two templates, we introduce a Kullback--Leibler (KL) divergence loss:
\begin{equation}
    \mathcal{L}_{\text{kl}} = \frac{1}{N_s} \sum_{i=1}^{N_s} \Bigl( \text{KL}(\hat{y}_i^t \| \hat{y}_i^{s}) + \text{KL}(\hat{y}_i^{s} \| \hat{y}_i^t) \Bigr)
\end{equation}
where \(N_s\) denotes the number of support samples per batch.

Finally, the standard cross-entropy loss is applied to the predicted labels:
\begin{equation}
    \mathcal{L}_{\text{ce}} = -\frac{1}{N_s} \sum_{i=1}^{N_s} \bigl[ y_i \log(\hat{y}_i) + (1 - y_i) \log(1 - \hat{y}_i) \bigr]
\end{equation}
with \(y_i\) representing the ground-truth label of each sample and \(N_s\) again being the batch size of the support set.

The overall training objective for the source-language stage is the weighted combination of the above components:
\begin{equation}
    \mathcal{L}_{\text{total}} = \mathcal{L}_{\text{ce}}^s + \mathcal{L}_{\text{ce}}^t + \mathcal{L}_{\text{kl}} + \alpha \cdot \mathcal{L}_c
\end{equation}
where \(\mathcal{L}_{\text{ce}}^s\) and \(\mathcal{L}_{\text{ce}}^t\) correspond to the cross-entropy losses computed from the source-language and cross-lingual templates, respectively, and \(\alpha\) is a hyperparameter that controls the contribution of the contrastive loss.

\subsection{Teacher-Student Distillation}

Having fine-tuned mBERT on the source language, the model has already acquired foundational knowledge for stance detection. The subsequent stage aims to adapt the model to the target-language environment using the available target-language resources. However, such resources are typically scarce, often an order of magnitude smaller than those in the source language. Consequently, we refrain from constructing separate support and query sets in this stage, as the reliable estimation of class prototypes would require a substantial number of samples that are unavailable in this low-resource setting.

To mitigate the challenge of limited data, we augment the target-language training corpus with inferential rationales generated by a Large Language Model (LLM). For the student model, we continue to employ the mBERT instance that was initialized and pre-fine-tuned in Source-Language Training step, thereby ensuring low computational cost and inference latency. In this stage, the LLM is used as a frozen rationale generator. The distillation is performed between two mBERT-based branches: a rationale-enhanced teacher branch that takes \((t_i, c_i, r_i)\) as input, where \(r_i\) denotes the rationale, and a standard student branch that takes only \((t_i, c_i)\) as input. In this way, the student branch learns from the reasoning-aware representations and predictions of the teacher branch while preserving rationale-free inference.

During training, the standard path processes the \((t_i, c_i)\) tuple through the encoder to produce the \([CLS]\) embedding of the final hidden layer, denoted as \(F_i\):
\begin{equation}
    F_i = \text{mBERT}((target_i, text_i))
\end{equation}
Analogously, the reasoning path processes the augmented tuple \((t_i, c_i, r_i)\) to yield the corresponding \([CLS]\) embedding \(F_{ri}\):
\begin{equation}
    F_{ri} = \text{mBERT}((target_i, text_i, reason_i))
\end{equation}

The embeddings \(F_i\) and \(F_{ri}\) are subsequently fed into a Contrastive Representation Distillation module, which serves dual purposes. From a contrastive perspective, it aligns the feature representations originating from the two different paths for the same sample while simultaneously pushing apart the representations of distinct samples. In terms of distillation, the module distills the rich representation information from the teacher branch to the student branch. This dual constraint facilitates effective knowledge distillation and enhances the model's discriminative capacity.

First, we compute the cosine similarity between embeddings from the standard path and those from the reasoning path:
\begin{equation}
    \text{sim}(i, j) = \frac{F_i \cdot F_{rj}}{\|F_i\| \, \|F_{rj}\|}
\end{equation}
where \(F_i\) denotes the embedding of the \(i\)-th sample from the standard path, and \(F_{rj}\) denotes the embedding of the \(j\)-th sample from the reasoning path.

Based on these similarities, we formulate a contrastive loss using the InfoNCE objective:
\begin{equation}
    \mathcal{L}_{c} = -\sum_{i=1}^{N} \log \frac{\exp(\text{sim}(i, i) / \tau)}{\sum_{j=1}^{N} \exp(\text{sim}(i, j) / \tau)}
\end{equation}
where \(\tau\) is a temperature hyperparameter and \(N\) is the batch size.

The embeddings from both paths are then passed through the classifier, which retains the same two-layer architecture employed in the Source-Language Training step:
\begin{equation}
    \hat{y}_i = \text{Classifier}(F_i)
\end{equation}
where \(\hat{y}_i\) represents the predicted stance probability. The same classifier is applied to the reasoning path embeddings \(F_{ri}\) in an identical manner.

Besides the representation-level distillation, we design a response-level distillation which transfers the soft label information from the teacher to the student, which assists with the representation distillation to achieve a more efficient knowledge transfer:
\begin{equation}
    \mathcal{L}_{\text{kl}} = \frac{1}{N_s} \sum_{i=1}^{N_s} \Bigl( \text{KL}(\hat{y}_i^t \| \hat{y}_i^{s}) + \text{KL}(\hat{y}_i^{s} \| \hat{y}_i^t) \Bigr)
\end{equation}
where \(\hat{y}_i^{t}\) and \(\hat{y}_i^s\) refer to the prediction of the teacher model and the student model, respectively.

Finally, the student model is supervised using the ground-truth stance labels via the cross-entropy loss:
\begin{equation}
    \mathcal{L}_{ce} = -\frac{1}{N} \sum_{i=1}^{N} \bigl[ y_i \log(\hat{y}_i^s) + (1 - y_i) \log(1 - \hat{y}_i^s) \bigr]
\end{equation}
where \(y_i\) is the ground-truth label of the \(i\)-th sample.

The overall training objective for the target-language stage is the weighted sum of the individual loss components:
\begin{equation}
    \mathcal{L}_{\text{total}} = \mathcal{L}_{ce} + \mathcal{L}_{kl} + \beta \cdot \mathcal{L}_c
\end{equation}
Here, \(\beta\) is a trade-off hyperparameter that balances the contribution of the contrastive constraint.

\section{Experiments and Results}

This section presents a comprehensive evaluation of our proposed method alongside detailed experimental configurations. We begin by introducing the datasets utilized in our study in Section \ref{datasets}, which include the X-stance, CIC, and VaxxStance datasets. Section \ref{baseline} delineates the comparative baselines, spanning both monolingual and cross-lingual stance detection approaches. In Section \ref{sec:main-exp}, we report the primary experimental results across all datasets, providing a performance comparison with baselines to verify the effectiveness of our framework. Furthermore, Section \ref{sec:reason-quality} investigates the impact of rationale quality on model performance by varying the number of in-context examples. In Section \ref{sec:support-query}, we justify the 4:1 ratio for constructing support and query sets in the source-language training phase through empirical analysis. The influence of various hyperparameters on experimental outcomes is discussed in Section \ref{sec:hyperparameters}. Additionally, we provide qualitative analysis in Section \ref{sec:qualitative-results} to offer deeper insights into the model's behavior. Finally, Section \ref{modes} explores the effects of different reasoning modes within the Qwen model on the overall performance.

\subsection{Dataset} \label{datasets}
(1) \textbf{X-stance} \cite{xstance}: X-stance is a large-scale multilingual and multi-target stance detection dataset focusing on Swiss politics. It encompasses comments in German, French, and Italian, where each sample consists of a candidate's response to a specific target question under various political themes (e.g., society, education, economy), labeled as either ``favor'' or ``against.'' In our experimental setup, we designate German as the source language and French as the target language. We select four primary topics: Society, Economy, Foreign Policy, and Education. The training set includes 3,430, 4,076, 2,521, and 4,604 samples in German for each respective topic, while approximately 500 French samples are provided for each as target-language data. Furthermore, to evaluate the cross-target generalization capability of our model, we ensure that each topic contains at least 15 distinct targets.\\
(2) \textbf{CIC} \cite{cic}: CIC is a multilingual stance detection dataset specifically curated to study the movement for Catalan independence within the linguistic context of Spanish and Catalan. It consists of social media posts (primarily from Twitter) where each entry represents a user's perspective on the ``Independence of Catalonia''. Each post is annotated with a stance label of ``favor'' or ``against''. In our settings, we utilize Spanish as the source language and Catalan as the target language, with the training set comprising 4,935 source samples and 510 target samples.\\
(3) \textbf{VaxxStance} \cite{vaxxstance}: VaxxStance is a multilingual dataset centered on the anti-vaccination debate. Each sample consists of a tweet directed at specific targets related to the vaccination movement (e.g., mandatory vaccination, pharmaceutical companies) within the domain of healthcare policy. The labels are categorized into ``favor'' and ``against.'' This dataset is particularly suitable for evaluating stance detection in low-resource and cross-lingual transfer scenarios. In our configuration, we adopt a cross-dataset transfer setting: the model is pre-trained on the German ``Society'' topic from X-stance (source) and then adapted to the Spanish ``vaccines'' target in VaxxStance (target), which contains 430 training samples.

\subsection{Comparative Baselines} \label{baseline}
To evaluate the effectiveness of our proposed method, we compare it against several state-of-the-art baselines, including both monolingual cross-target and specialized cross-lingual stance detection frameworks:

\textbf{TAN} (Target-specific Attention Network) \cite{tan} is a classic stance detection framework that utilizes an attention mechanism to capture semantic interactions between a post and its target. By prioritizing target-relevant information, TAN effectively identifies stance-bearing expressions to improve classification accuracy.

\textbf{CrossNet} \cite{crossnet} is a self-attention-based model specifically developed for cross-target stance detection. It captures latent connections and shared linguistic patterns between a source and a destination target, enabling the transfer of stance-related knowledge to unseen targets.

\textbf{JointCL} \cite{jointcl} is a framework for zero-shot stance detection that integrates stance contrastive learning with target-aware prototypical graph contrastive learning. It generalizes stance representations to unseen targets by leveraging prototypical graphs to share reasoning patterns across different domains.

\textbf{CLKD} (Cross-Lingual Knowledge Distillation) \cite{clkd} is a cross-lingual distillation framework that trains classifiers for unlabeled target languages. It uses soft predictions from a source-language teacher model as supervision on parallel corpora, combining distillation with adversarial adaptation to bridge the language gap.

\textbf{mBERT-FT} \cite{bert} serves as a fundamental baseline where mBERT is directly fine-tuned on the target-language training data. Unlike our method, this baseline does not incorporate dual training paths or specialized stance detection templates.

\textbf{CCSD} (Cross-Lingual Cross-Target Stance Detection) \cite{zhang2023cross} is a framework designed to address both language barriers and target inconsistency by transferring linguistic and target-oriented knowledge. The original model employs dual knowledge distillation framework consisting of a cross-lingual teacher and a cross-target teacher. To ensure a fair comparison within our few-shot experimental setting, we adapt the CCSD framework by providing a small amount of labeled target-language data to its student model for training, aligning it with the data availability of our proposed method.

\subsection{Results and Analysis} \label{sec:main-exp}

\begin{table}[htbp]
\centering
\caption{Performance comparison between our method and competitive baselines on the X-stance, CIC, and VaxxStance datasets. Accuracy (Acc) and macro \(F_1\)-score (\(F_1\)) are reported. ``improve (vs. CCSD)'' denotes the absolute performance gain of our best setting over the strongest baseline CCSD.}
\label{results}
\small
\setlength{\tabcolsep}{3pt}
\makebox[\textwidth][c]{
    \begin{tabular}{llcccccccccccc}
    \toprule
    \multicolumn{2}{l}{\multirow{2}{*}{method}} & \multicolumn{2}{c}{X-Society} & \multicolumn{2}{c}{X-Economy} & \multicolumn{2}{c}{X-Education} & \multicolumn{2}{c}{X-Policy} & \multicolumn{2}{c}{CIC} & \multicolumn{2}{c}{VaxxStance} \\ \cmidrule(lr){3-4} \cmidrule(lr){5-6} \cmidrule(lr){7-8} \cmidrule(lr){9-10} \cmidrule(lr){11-12} \cmidrule(lr){13-14}
    \multicolumn{2}{l}{} & Acc & \(F_1\) & Acc & \(F_1\) & Acc & \(F_1\) & Acc & \(F_1\) & Acc & \(F_1\) & Acc & \(F_1\) \\ \midrule
    \multicolumn{2}{l}{TAN \cite{tan}} & 61.84 & 60.91 & 60.84 & 59.91 & 63.95 & 63.21 & 63.26 & 62.84 & 61.89 & 60.70 & 64.82 & 62.17 \\
    \multicolumn{2}{l}{CrossNet \cite{crossnet}} & 63.35 & 63.09 & 62.70 & 61.94 & 64.82 & 63.90 & 64.80 & 63.51 & 61.57 & 60.31 & 67.92 & 65.91 \\
    \multicolumn{2}{l}{JointCL \cite{jointcl}} & 64.64 & 64.18 & 64.22 & 64.03 & 66.56 & 66.37 & 67.05 & 66.53 & 64.98 & 63.59 & 72.81 & 70.79 \\ \midrule
    \multicolumn{2}{l}{CLKD \cite{clkd}} & 66.81 & 65.94 & 67.10 & 66.31 & 69.84 & 69.06 & 70.41 & 68.24 & 69.25 & 67.82 & 73.92 & 71.98 \\
    \multicolumn{2}{l}{mBERT-FT \cite{bert}} & 66.89 & 66.27 & 66.47 & 66.15 & 70.04 & 69.63 & 70.62 & 70.20 & 68.82 & 66.14 & 76.81 & 74.82 \\
    \multicolumn{2}{l}{CCSD \cite{zhang2023cross}} & 68.92 & 67.91 & 68.26 & 67.81 & 72.94 & 71.86 & 73.26 & 72.96 & 68.90 & 66.23 & \underline{77.86} & \underline{75.83} \\ \midrule
    \multirow{2}{*}{\textbf{ours}} & Qwen & \underline{71.02} & \underline{69.83} & \underline{69.83} & \underline{69.65} & \underline{74.12} & \underline{73.20} & \underline{74.29} & \underline{73.53} & \underline{71.85} & \textbf{70.14} & 77.82 & 75.51 \\
     & Qwen thinking & \textbf{71.54} & \textbf{70.05} & \textbf{70.59} & \textbf{70.14} & \textbf{74.29} & \textbf{73.77} & \textbf{74.35} & \textbf{73.69} & \textbf{72.24} & \underline{70.08} & \textbf{79.23} & \textbf{75.95} \\ \midrule
    \multicolumn{2}{l}{improve (vs. CCSD)} & \textit{\textcolor{red}{+2.62}} & \textit{\textcolor{red}{+2.14}} & \textit{\textcolor{red}{+2.33}} & \textit{\textcolor{red}{+2.33}} & \textit{\textcolor{red}{+1.35}} & \textit{\textcolor{red}{+1.91}} & \textit{\textcolor{red}{+1.09}} & \textit{\textcolor{red}{+0.73}} & \textit{\textcolor{red}{+2.34}} & \textit{\textcolor{red}{+3.91}} & \textit{\textcolor{red}{+1.37}} & \textit{\textcolor{red}{+0.12}} \\ \bottomrule
    \end{tabular}
    }
\end{table}

We evaluate our proposed framework across the X-stance, CIC, and VaxxStance datasets, covering six distinct topics. Performance is measured using Accuracy and the Macro \(F_1\)-score as the primary evaluation metrics. The LLM utilized in our approach is Qwen 3.5 Flash  \cite{qwen3.5}, which is tested in both non-thinking and thinking modes. To demonstrate the superiority of our method, we compare its performance against several competitive monolingual and cross-lingual stance detection baselines.

Among the monolingual baselines, the prototypical graph mechanism in JointCL \cite{jointcl} bridges the gap between known and unknown targets more effectively than attention-based architectures like TAN \cite{tan} and CrossNet \cite{crossnet}. This advantage suggests that extracting high-level semantic features is crucial for mitigating target inconsistency. In cross-lingual scenarios, direct fine-tuning of mBERT yields competitive results, leveraging its robust pre-trained multilingual representations. Similarly, CLKD \cite{clkd} exhibits strong performance, validating that cross-lingual knowledge distillation is a highly viable strategy in the absence of target-language labels. Furthermore, the CCSD \cite{zhang2023cross} framework, which also employs mBERT as its backbone, serves as a strong baseline for joint cross-lingual and cross-target tasks.

The comprehensive experimental results are summarized in Table~\ref{results}. Our method consistently outperforms all baselines, achieving a performance gain of approximately 2\% to 3\% over the state-of-the-art CCSD framework. Specifically, on the X-stance dataset, our method demonstrates significant improvements in the Society and Economy topics, with an average increase of 2.48\% in Accuracy and 2.24\% in \(F_1\)-score. On the CIC dataset, our approach substantially exceeds the CCSD baseline, particularly in terms of \(F_1\)-score, which sees a marked improvement of 3.91\%. In the VaxxStance dataset, we observe a steady improvement of 1.37\% in Accuracy over CCSD.

Compared with previous approaches, our framework derives more robust representations by constructing class prototypes and optimizing the model via calculating the distance between query samples and class centroids. This prototypical approach mitigates the impact of individual sample noise on the overall representation. Additionally, by incorporating LLM-generated rationales to augment the target-language data, we guide mBERT to focus on critical linguistic features and emulate the systematic reasoning process of the LLM. In the following sections, we provide a detailed ablation study to further verify the effectiveness of the individual components within our framework.

\subsection{The Effect of LLM-Generated Rationales} \label{sec:reason-exp}
\begin{figure}[tbp]
\centering
\subfloat[\centering]{\includegraphics[width=7.0cm]{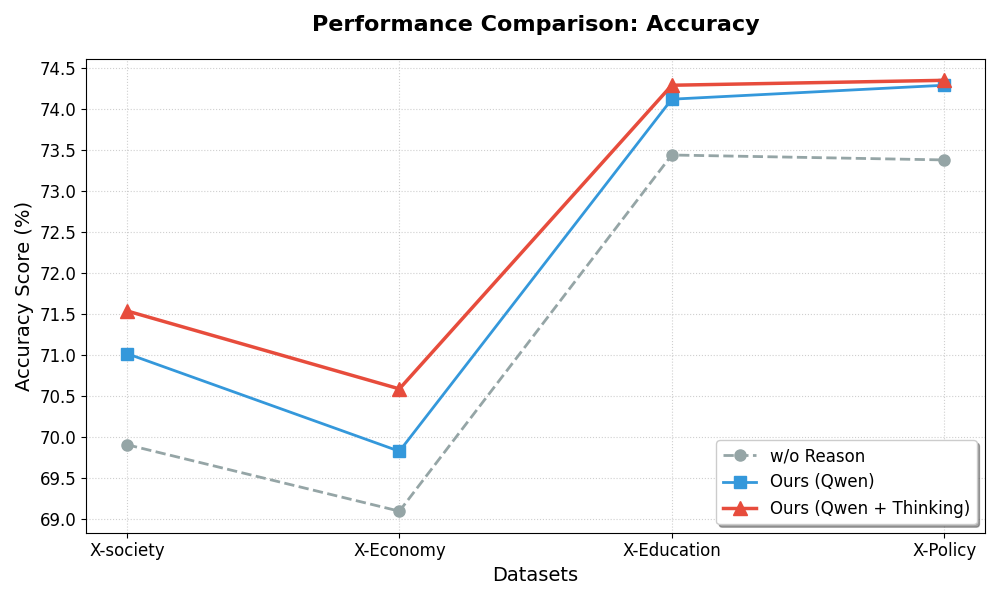}}
\subfloat[\centering]{\includegraphics[width=7.0cm]{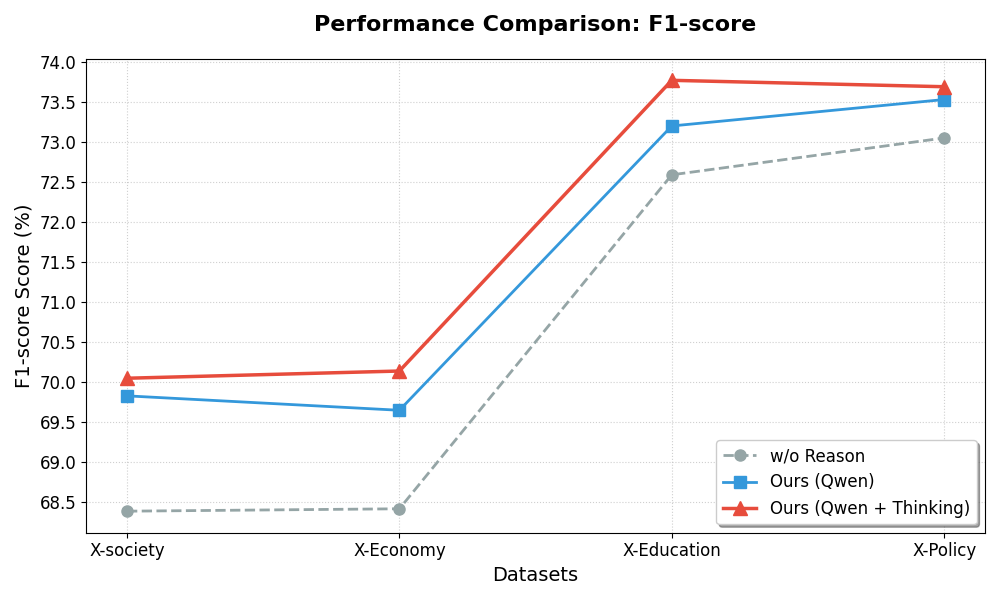}}
\caption{Effectiveness of the LLM reasoning on four X-stance topics. \textbf{(a)} Accuracy comparison of w/o Rationales, Ours (Qwen), and Ours (Qwen + Thinking). \textbf{(b)} \(F_1\)-score comparison of the same three settings.}
\Description{Accuracy and \(F_1\) comparisons showing the effect of LLM-generated rationales across X-stance topics.}
\label{fig:reasoning-wo}
\end{figure}

One of the core designs of our framework is the reasoning path integrated into the Teacher-Student Distillation stage. In this section, we investigate the empirical effectiveness of incorporating LLM-generated rationales into our method.

To evaluate the contribution of these rationales, we conducted ablation experiments by removing the reasoning path in the distillation step, thereby retaining only the standard training path. These experiments were performed across the X-Society, X-Economy, X-Education, and X-Policy datasets, with the comparative results illustrated in Fig~\ref{fig:reasoning-wo}. As indicated by the gray line in the figure which represents the framework performance in the absence of reasoning-enhanced data, the exclusion of rationales leads to a noticeable decline in overall performance. Specifically, compared to the framework utilizing Qwen in non-thinking mode, the Accuracy and \(F_1\)-score drop by approximately 1.0\% and 1.5\%, respectively.

A more granular analysis reveals that the performance degradation is most pronounced in the X-Society dataset, where Accuracy decreases by 1.11\% compared to the Qwen non-thinking mode and by 1.63\% compared to the thinking mode. Similarly, the \(F_1\)-score in this dataset falls by 1.44\% and 1.66\% relative to the two LLM modes. In contrast, the X-Education dataset exhibits a more moderate decline, with Accuracy and \(F_1\)-score dropping by 0.68\% and 0.61\%, respectively. These results consistently demonstrate that the removal of LLM rationales leads to a significant reduction in model efficacy, thereby validating the critical role and effectiveness of the reasoning path in our target-language framework.

\subsection{The Effect of Rationale Quality} \label{sec:reason-quality}

\begin{table}[tbp]
\centering
\caption{Ablation study on the number of ICL demonstrations used for rationale generation with Qwen thinking mode. The numbers in parentheses denote the absolute change relative to the 8-sample setting, showing that 8 demonstrations provide the best overall trade-off between effectiveness and prompt length.}
\label{icl_samples_performance}
\small
\begin{tabular}{lcccc}
\toprule
\multirow{2}{*}{Setting} & \multicolumn{2}{c}{X-Society} & \multicolumn{2}{c}{X-Economy} \\
\cmidrule(lr){2-3} \cmidrule(lr){4-5}
 & Acc ($\Delta$) & \(F_1\) ($\Delta$) & Acc ($\Delta$) & \(F_1\) ($\Delta$) \\ \midrule
2-samples & 70.62 (\textcolor{blue}{-0.85}) & 69.24 (\textcolor{blue}{-0.81}) & 69.54 (\textcolor{blue}{-1.05}) & 69.18 (\textcolor{blue}{-0.96}) \\
4-samples & 71.10 (\textcolor{blue}{-0.37}) & 69.73 (\textcolor{blue}{-0.32}) & 70.06 (\textcolor{blue}{-0.53}) & 69.62 (\textcolor{blue}{-0.52}) \\
6-samples & 71.25 (\textcolor{blue}{-0.29}) & 69.82 (\textcolor{blue}{-0.23}) & 70.35 (\textcolor{blue}{-0.24}) & 69.91 (\textcolor{blue}{-0.23}) \\
\textbf{8-samples (ours)} & \textbf{71.54}  & \textbf{70.05}   & \textbf{70.59}   & \textbf{70.14}   \\
10-samples & 71.47 (\textcolor{blue}{-0.07}) & 70.07 (\textcolor{red}{+0.02}) & 70.62 (\textcolor{red}{+0.03}) & 70.21 (\textcolor{red}{+0.07}) \\
\midrule \midrule
\multirow{2}{*}{Setting} & \multicolumn{2}{c}{X-Education} & \multicolumn{2}{c}{X-Policy} \\
\cmidrule(lr){2-3} \cmidrule(lr){4-5}
 & Acc ($\Delta$) & \(F_1\) ($\Delta$) & Acc ($\Delta$) & \(F_1\) ($\Delta$) \\ \midrule
2-samples & 73.75 (\textcolor{blue}{-0.54}) & 73.37 (\textcolor{blue}{-0.40}) & 73.74 (\textcolor{blue}{-0.61}) & 73.29 (\textcolor{blue}{-0.40}) \\
4-samples & 73.97 (\textcolor{blue}{-0.32}) & 73.52 (\textcolor{blue}{-0.25}) & 74.09 (\textcolor{blue}{-0.26}) & 73.41 (\textcolor{blue}{-0.28}) \\
6-samples & 74.18 (\textcolor{blue}{-0.11}) & 73.71 (\textcolor{blue}{-0.06}) & 74.18 (\textcolor{blue}{-0.17}) & 73.52 (\textcolor{blue}{-0.17}) \\
\textbf{8-samples (ours)} & \textbf{74.29}   & \textbf{73.77}   & \textbf{74.35}   & \textbf{73.69}   \\
10-samples & 74.54 (\textcolor{red}{+0.25}) & 73.91 (\textcolor{red}{+0.14}) & 74.31 (\textcolor{blue}{-0.04}) & 73.72 (\textcolor{red}{+0.03}) \\ \bottomrule
\end{tabular}
\end{table}

In the rationale preparation stage, we employ In-Context Learning (ICL) to generate high-quality rationales rich in inferential information, following the Chain-of-Thought paradigm: ``Let's think step by step. First, the word in the text indicates that... Second, ...''. In this section, we investigate the optimal number of in-context examples required to achieve this effect while balancing computational efficiency. We conducted five sets of experiments, comparing 2, 4, 6, 8, and 10 examples using the Qwen thinking mode across the X-Society, X-Education, X-Economy, and X-Policy datasets. The results are summarized in Table~\ref{icl_samples_performance}.

As shown in the table, model performance generally improves as the number of in-context examples increases, reaching a plateau at approximately 8 samples. In the 2-sample configuration, the performance across the four datasets is relatively suboptimal; for instance, the Accuracy in X-Society is 70.62\%, which is 0.85\% lower than that of the 8-sample setting. Increasing the number of examples to four yields substantial gains, with Accuracy and \(F_1\)-score in X-Society improving by 0.48\% and 0.49\%, respectively. While performance continues to grow when expanding to 6 samples, the marginal gains begin to diminish. For example, the improvement in X-Society is only 0.09\% for both Accuracy and \(F_1\)-score, representing a significant slowdown in the growth rate compared to the previous increment.

Similar trends are observed in other datasets. At 8 samples, the performance increment remains positive but slight. In the X-Economy and X-Education datasets, the growth in Accuracy further tapers to 0.24\% and 0.11\%, respectively, suggesting that the model is approaching its performance ceiling. Upon increasing the count to 10 samples, the performance stagnates across most datasets. Specifically, Accuracy in X-Society even experiences a slight decrease of 0.07\%, while the improvements in X-Economy and X-Policy are negligible (less than 0.07\%). Although the 10-sample setting occasionally offers marginal benefits, we ultimately selected the 8-sample configuration. This decision strikes an optimal balance between effectiveness and efficiency, as utilizing 8 examples reduces input token consumption by approximately 15\% compared to the 10-sample setting, significantly lowering the overall computational cost.

\subsection{The Effect of Support-Query Set Construction}\label{sec:support-query}

\begin{table}[tbp]
\centering
\caption{Effect of different support-query set constructions in source-language training. Accuracy (Acc) and \(F_1\)-score (\(F_1\)) are reported on the four X-stance topics under different combinations of support and query samples. The values in parentheses indicate the performance difference relative to the default setting of 8 support samples per class + 4 query samples, which provides the best trade-off between performance and computational cost.}
\label{performance_stacking_v2}
\small
\begin{tabular}{lcccc}
\toprule
\multirow{2}{*}{Setting} & \multicolumn{2}{c}{X-Society} & \multicolumn{2}{c}{X-Economy} \\
\cmidrule(lr){2-3} \cmidrule(lr){4-5}
 & Acc ($\Delta$) & \(F_1\) ($\Delta$) & Acc ($\Delta$) & \(F_1\) ($\Delta$) \\ \midrule
4 support × 2 + 1 query  & 69.93 (\textcolor{blue}{-1.61}) & 69.44 (\textcolor{blue}{-0.61}) & 70.26 (\textcolor{blue}{-0.33}) & 69.62 (\textcolor{blue}{-0.52}) \\
4 support × 2 + 2 query & 71.02 (\textcolor{blue}{-0.52}) & 69.52 (\textcolor{blue}{-0.53}) & 70.21 (\textcolor{blue}{-0.38}) & 69.64 (\textcolor{blue}{-0.50}) \\
4 support × 2 + 4 query & 71.18 (\textcolor{blue}{-0.36}) & 69.75 (\textcolor{blue}{-0.30}) & 70.38 (\textcolor{blue}{-0.21}) & 69.86 (\textcolor{blue}{-0.28}) \\
\textbf{8 support × 2 + 4 query (ours)} & \textbf{71.54}   & \textbf{70.05}   & \textbf{70.59}   & \textbf{70.14}   \\
10 support × 2 + 4 query & 71.63 (\textcolor{red}{+0.09}) & 70.09 (\textcolor{red}{+0.04}) & 70.63 (\textcolor{red}{+0.04}) & 70.07 (\textcolor{blue}{-0.07}) \\
\midrule \midrule
\multirow{2}{*}{Setting} & \multicolumn{2}{c}{X-Education} & \multicolumn{2}{c}{X-Policy} \\
\cmidrule(lr){2-3} \cmidrule(lr){4-5}
 & Acc ($\Delta$) & \(F_1\) ($\Delta$) & Acc ($\Delta$) & \(F_1\) ($\Delta$) \\ \midrule
4 support × 2 + 1 query & 73.94 (\textcolor{blue}{-0.35}) & 73.40 (\textcolor{blue}{-0.37}) & 74.06 (\textcolor{blue}{-0.29}) & 73.31 (\textcolor{blue}{-0.38}) \\
4 support × 2 + 2 query & 74.08 (\textcolor{blue}{-0.21}) & 73.48 (\textcolor{blue}{-0.29}) & 74.15 (\textcolor{blue}{-0.20}) & 73.42 (\textcolor{blue}{-0.27}) \\
4 support × 2 + 4 query & 74.07 (\textcolor{blue}{-0.22}) & 73.51 (\textcolor{blue}{-0.26}) & 74.13 (\textcolor{blue}{-0.22}) & 73.50 (\textcolor{blue}{-0.19}) \\
\textbf{8 support × 2 + 4 query (ours)} & \textbf{74.29}   & \textbf{73.77}   & \textbf{74.35}   & \textbf{73.69}   \\
10 support × 2 + 4 query & 74.37 (\textcolor{red}{+0.08}) & 73.90 (\textcolor{red}{+0.13}) & 74.35 (0.00) & 73.74 (\textcolor{red}{+0.05}) \\ \bottomrule
\end{tabular}
\end{table}

The construction of the support and query sets within the Source-Language Training path is pivotal, as different configurations can significantly influence model performance. In this section, we investigate various combinations of these sets to identify the optimal pattern for cross-lingual stance detection. Specifically, we vary the number of support samples per class from 4 to 10 to determine the minimum required to construct robust class prototypes. Concurrently, we evaluate query set sizes ranging from 1 to 4 samples to balance effective model optimization with computational efficiency.

Initially, we evaluated a configuration consisting of 4 support samples per class and a single query sample (maintaining a 4:1 ratio). However, the performance was suboptimal; for instance, in the X-Society dataset, both Accuracy and \(F_1\)-score remained below 70\%. This deficiency likely stems from the limited number of query samples, where individual noisy samples can disproportionately misguide the model's optimization. Increasing the query size to 2 samples yielded a measurable improvement, with Accuracy and \(F_1\)-score in the X-Society dataset rising by 1.09\% and 0.08\%, respectively. Similar gains across other datasets validate our hypothesis that a larger query set mitigates the impact of stochastic noise. While further increasing the query size to 4 samples continued to enhance performance, the marginal utility began to decrease, with negligible improvements observed in the X-Education and X-Policy datasets.

We speculate that when the support set is limited to 4 samples per class, the representative power of the class prototypes becomes the bottleneck. To address this, we increased the support set size to 8 samples per class, which resulted in a consistent performance boost of approximately 0.2\%--0.3\% across nearly all datasets. This indicates that a more substantial support set is essential for constructing robust and stable class prototypes. Although further expanding the support set to 10 samples per class provided additional gains, the improvement was minor -- less than 0.1\% in Accuracy across all datasets while significantly increasing computational overhead. Consequently, to achieve an optimal balance between empirical performance and computational cost, we designated a configuration of 8 support samples per class and 4 query samples as our standard setting.

\subsection{The Effects of Hyperparameters}\label{sec:hyperparameters}

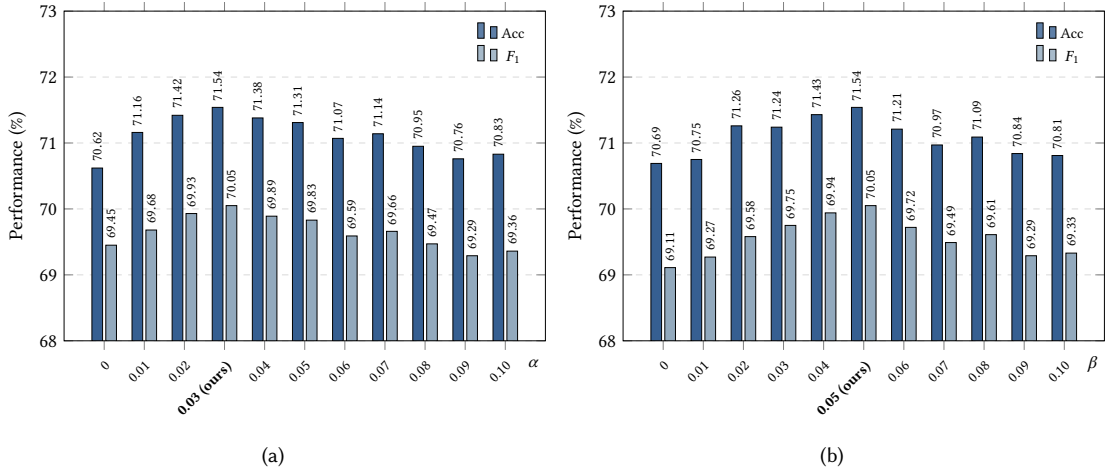
\begin{figure}[tbp]
            \centering
        \subfloat[\centering]{
            \resizebox{0.48\textwidth}{!}{%
            \begin{tikzpicture}[scale=1]
                \begin{axis}[
                    width=9.5cm, height=7cm,
                    ybar=1.5pt, bar width=5pt,
                    enlarge x limits=0.1,
                    xlabel={$\alpha$}, ylabel={Performance (\%)},
                    xlabel style={
                        at={(axis description cs:0.95,-0.07)},
                        anchor=west,
                        font=\small
                    },
                    symbolic x coords={0, 0.01, 0.02, 0.03, 0.04, 0.05, 0.06, 0.07, 0.08, 0.09, 0.10},
                    xtick=data,
                    xticklabels={0, 0.01, 0.02, \textbf{0.03 (ours)}, 0.04, 0.05, 0.06, 0.07, 0.08, 0.09, 0.10},
                    xticklabel style={rotate=45, anchor=north east, font=\footnotesize},
                    ymin=68, ymax=73,
                    ymajorgrids=true, grid style={dashed, gray!30},
                    legend style={
                        at={(0.98,0.98)},
                        anchor=north east,
                        draw=none,
                        fill=white,
                        fill opacity=0.8,
                        text opacity=1,
                        font=\footnotesize
                    },
                    legend columns=1,
                    nodes near coords,
                    every node near coord/.append style={font=\scriptsize, rotate=90, anchor=west, /pgf/number format/precision=2}
                ]
                    \addplot[fill=colorAcc] coordinates {
                        (0, 70.62) (0.01, 71.16) (0.02, 71.42) (0.03, 71.54) (0.04, 71.38)
                        (0.05, 71.31) (0.06, 71.07) (0.07, 71.14) (0.08, 70.95) (0.09, 70.76) (0.10, 70.83)
                    };
                    \addlegendentry{Acc}
                    \addplot[fill=colorF1] coordinates {
                        (0, 69.45) (0.01, 69.68) (0.02, 69.93) (0.03, 70.05) (0.04, 69.89)
                        (0.05, 69.83) (0.06, 69.59) (0.07, 69.66) (0.08, 69.47) (0.09, 69.29) (0.10, 69.36)
                    };
                    \addlegendentry{\(F_1\)}
                \end{axis}
            \end{tikzpicture}
            }
        }
        \hspace{-5pt}
        \subfloat[\centering]{
            \resizebox{0.48\textwidth}{!}{%
            \begin{tikzpicture}[scale=1]
                \begin{axis}[
                    width=9.5cm, height=7cm,
                    ybar=1.5pt, bar width=5pt,
                    enlarge x limits=0.1,
                    xlabel={$\beta$}, ylabel={Performance (\%)},
                    xlabel style={
                        at={(axis description cs:0.95,-0.07)},
                        anchor=west,
                        font=\small
                    },
                    symbolic x coords={0, 0.01, 0.02, 0.03, 0.04, 0.05, 0.06, 0.07, 0.08, 0.09, 0.10},
                    xtick=data,
                    xticklabels={0, 0.01, 0.02, 0.03, 0.04, \textbf{0.05 (ours)}, 0.06, 0.07, 0.08, 0.09, 0.10},
                    xticklabel style={rotate=45, anchor=north east, font=\footnotesize},
                    ymin=68, ymax=73,
                    ymajorgrids=true, grid style={dashed, gray!30},
                    legend style={
                        at={(0.98,0.98)},
                        anchor=north east,
                        draw=none,
                        fill=white,
                        fill opacity=0.8,
                        text opacity=1,
                        font=\footnotesize
                    },
                    legend columns=1,
                    nodes near coords,
                    every node near coord/.append style={font=\scriptsize, rotate=90, anchor=west, /pgf/number format/precision=2}
                ]
                    \addplot[fill=colorAcc] coordinates {
                        (0, 70.69) (0.01, 70.75) (0.02, 71.26) (0.03, 71.24) (0.04, 71.43)
                        (0.05, 71.54) (0.06, 71.21) (0.07, 70.97) (0.08, 71.09) (0.09, 70.84) (0.10, 70.81)
                    };
                    \addlegendentry{Acc}
                    \addplot[fill=colorF1] coordinates {
                        (0, 69.11) (0.01, 69.27) (0.02, 69.58) (0.03, 69.75) (0.04, 69.94)
                        (0.05, 70.05) (0.06, 69.72) (0.07, 69.49) (0.08, 69.61) (0.09, 69.29) (0.10, 69.33)
                    };
                    \addlegendentry{\(F_1\)}
                \end{axis}
            \end{tikzpicture}
            }
        }
    
    \caption{Effect of the contrastive loss weight on model performance. \textbf{(a)} Performance variation in the source-language training stage under different weights of the query-centered contrastive loss. \textbf{(b)} Performance variation in the teacher-student distillation stage under different weights of the contrastive representation distillation loss. In both subfigures, Accuracy (Acc) and \(F_1\)-score (\(F_1\)) are reported, and the best-performing setting is highlighted.}
    \Description{Bar charts showing performance sensitivity to the alpha and beta contrastive loss weights.}
    \label{fig:hyperparameters}
\end{figure}

Our proposed framework involves two key hyperparameters: $\alpha$ and $\beta$, which represent the weights of the contrastive losses in the Source-Language and Teacher-Student Distillation stages, respectively. All sensitivity analyses were conducted on the X-Society dataset utilizing the thinking mode of Qwen, with results illustrated in Fig~\ref{fig:hyperparameters}.

To evaluate the impact of the source-language contrastive loss, we vary $\alpha$ from 0 to 0.10 with an interval of 0.01. As shown in the bar chart, the Accuracy increases from 70.62\% at $\alpha=0$ to a peak of 71.54\% at $\alpha=0.03$. Beyond this threshold, the performance exhibits a downward trend, with Accuracy decreasing to 70.83\% as $\alpha$ reaches 0.10. The \(F_1\)-score follows a similar distribution, suggesting that a moderate contrastive penalty effectively regularizes the source representation.

Then, we examine the contrastive representation distillation loss weight $\beta$ in the target language within a range of 0 to 0.10 with an interval of 0.01. The results demonstrate that Accuracy reaches its optimum at $\beta=0.05$ and diminishes when the value deviates from this center. The \(F_1\)-score consistently follows this pattern, confirming that appropriate contrastive alignment in the target language is essential for robust cross-lingual transfer.

\subsection{Qualitative Results}\label{sec:qualitative-results}

\begin{figure}[tbp]

\includegraphics[width=0.85\textwidth]{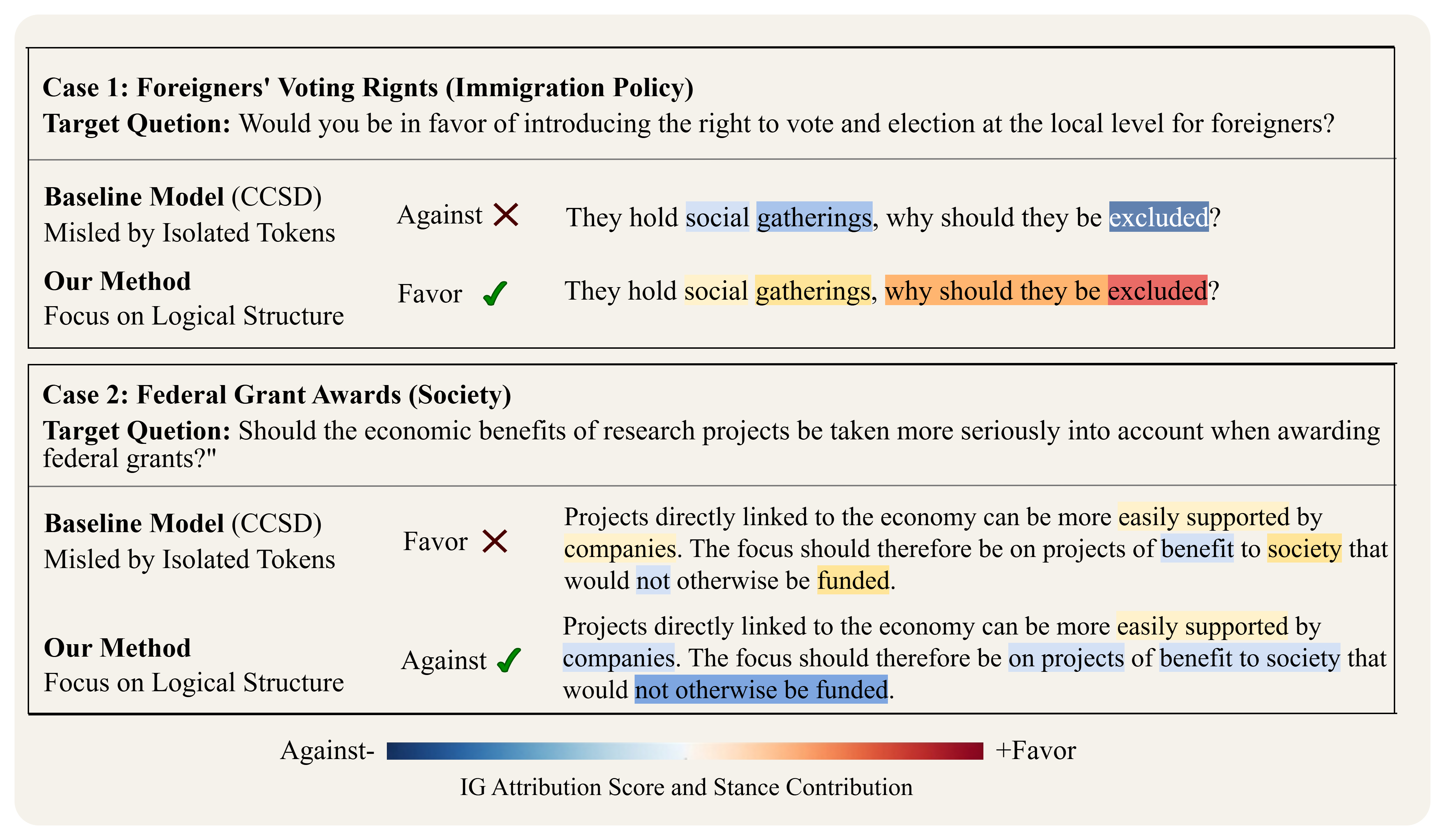}
\caption{Qualitative comparison between the baseline model and our method using Integrated Gradients (IG). Red indicates positive contribution toward Favor, and blue indicates positive contribution toward Against; darker colors denote stronger attribution scores.}
\Description{Integrated Gradients visualization comparing token-level stance attribution for the baseline model and the proposed method.}
\label{fig:qualitative}
\end{figure}

To provide an intuitive demonstration of our framework's effectiveness, we conducted a qualitative analysis using Integrated Gradients (IG) \cite{sundararajan2017axiomatic}, a gradient-based attribution method that assigns importance scores to each input token. In our visualization, red highlights denote positive contributions toward a ``Favor'' stance, while blue highlights indicate contributions toward ``Against''. The color intensity reflects the model's confidence and the magnitude of the token's impact on the final decision.

The qualitative results are illustrated in Fig~\ref{fig:qualitative}. In case 1, we detect the stance of the text \textit{``They hold social gatherings, why should they be excluded?''} toward the target \textit{`` Would you be in favor of introducing the right to vote and election at the local level for foreigners?''}. Our method identifies the phrase \textit{``why should they be excluded''} as a whole to identify the coherent favor stance, while the baseline model is misled by the single word \textit{``excluded''}. In case 2, our model demonstrates its proficiency in integrating contextual information from entire phrases rather than relying on isolated keywords, leading to more robust and trustworthy stance detection.

\subsection{The Effect of LLM Reasoning Modes}\label{modes}

By comparing the performance of the Qwen thinking mode and non-thinking mode within our framework, we observe that the thinking mode consistently outperforms the non-thinking mode across nearly all dataset configurations. Our empirical analysis suggests that this superiority is primarily attributed to the enhanced quality and higher degree of cleanliness in the generated data. Through manual qualitative inspection of the rationales produced by both modes, we find that the thinking mode yields responses that are typically more accurate and characterized by greater logical depth. In contrast, rationales generated in the non-thinking mode occasionally exhibit instability, manifesting as structural inconsistencies, noisy characters, or irrelevant information. Such artifacts can introduce undesirable noise into the training process, potentially misleading mBERT and undermining its accuracy in stance detection tasks. Consequently, the advanced reasoning capabilities inherent in the thinking mode provide more robust and reliable supervisory signals for cross-lingual knowledge transfer.

\subsection{Implementation Details}\label{sec:implementation-details}

All models are optimized using Adam with a learning rate of \(2 \times 10^{-5}\), and the number of training epochs is set to 15. We use multilingual BERT (mBERT) as the student backbone, while rationale generation is performed by Qwen3.5-Flash \cite{qwen3.5} which supports a maximum context window of 1,000,000 tokens and a maximum output length of 64,000 tokens. Specifically, we access the hosted API version of Qwen3.5-Flash, rather than deploying the model locally, and keep the LLM parameters frozen throughout the rationale generation stage. We evaluate both the thinking and non-thinking modes of Qwen3.5-Flash to analyze the influence of rationale quality. Based on the hyperparameter analysis above, the contrastive loss weight in source-language training is set to \(\alpha=0.03\), and the contrastive representation distillation weight in target-language training is set to \(\beta=0.05\). For rationale generation, we use 8 in-context examples, which provide the best trade-off between effectiveness and prompt length in our experiments.

\section{Discussion}
The experimental results presented in this paper provide strong support for our hypothesis that leveraging the reasoning capabilities of Large Language Models through knowledge distillation can significantly improve cross-lingual stance detection, especially in low-resource settings. Our framework consistently outperforms state-of-the-art baselines such as CCSD, CLKD, and JointCL across multiple multilingual datasets (X-stance, CIC, and VaxxStance), with accuracy gains ranging from approximately 1\% to 3\%. Such results align with the observations of previous studies that emphasize the importance of logical reasoning over superficial semantic correlations. Our work demonstrates that incorporating explicit Chain-of-Thought rationales generated by a frozen LLM provides richer supervisory signals that help the student model (mBERT) to focus on critical linguistic cues and avoid being misled by isolated keywords, as illustrated in our qualitative analysis.

Our dual-branch hierarchical distillation framework is designed to overcome the architectural mismatch between decoder-only Large Language Models and encoder-only mBERT. The consistent performance gains validate that aligning both representation-level and response-level outputs across the reasoning and standard paths enables effective knowledge transfer without directly aligning LLM outputs with mBERT outputs. Furthermore, the introduction of a contrastive learning mechanism for both the source language and the target language proves essential. Our ablation study shows that removing the reasoning path leads to a noticeable decline of approximately 1.0\%--1.5\% in accuracy and \(F_1\), confirming that the rationales are central to the model's improved discriminative capacity. The sensitivity analysis on hyperparameters \(\alpha\) and \(\beta\) further reveals that moderate contrastive penalties yield optimal results, echoing findings in prior contrastive learning literature that excessive constraints can hurt generalization.

The broader implications of our findings extend beyond stance detection. In many real-world applications such as public opinion mining, market analysis, and content recommendation systems, low-resource languages are often neglected due to data scarcity. Our framework offers a practical pathway to transfer reasoning-rich knowledge from high-resource to low-resource languages without requiring large-scale parallel corpora or expensive LLM deployment at inference time. The use of Qwen in thinking mode, which produces cleaner and more logically coherent rationales, underscores the importance of reasoning quality; this suggests that even moderately sized Large Language Models can serve as effective teachers when properly prompted with few-shot examples. The optimal selection of 8 in-context examples (Section \ref{sec:reason-quality}) strikes a practical balance between performance and computational cost, reducing token consumption by 15\% compared to 10 examples while maintaining near-peak accuracy.

Nevertheless, several limitations and future directions should be noted. First, our method still relies on a small amount of labeled target-language data (approximately 500 samples per topic). Extending to a truly zero-shot setting where no target-language labels are available would require additional mechanisms such as unsupervised domain adaptation or more sophisticated prompt engineering. Second, while we focus on encoder-only mBERT for efficiency, exploring other compact architectures (e.g., DistilBERT or XLM-RoBERTa) could further improve deployment flexibility. Third, the reasoning rationales are generated by a single LLM (Qwen); using LLM ensembles or iterative rationale refinement might yield even higher quality. Finally, our current framework assumes binary stance labels (favor/against); extending to multi-class or fine-grained stance detection is a promising avenue. Future work could also investigate dynamic adjustment of the number of in-context examples based on topic complexity, as well as incorporating external knowledge bases to further enrich the reasoning process without increasing inference latency.

\section{Conclusions}

In this paper, we proposed a novel knowledge distillation framework that leverages the reasoning capabilities of Large Language Models to improve cross-lingual stance detection, particularly for low-resource languages. Our approach addresses two fundamental challenges: the architectural mismatch between decoder-only Large Language Models and encoder-only mBERT, and the scarcity of labeled data in target languages. To overcome these issues, we designed a dual-branch hierarchical distillation mechanism that transfers both representation-level and response-level knowledge from LLM-generated rationales to the student model. Additionally, we introduced two complementary contrastive learning strategies: prototype-based contrastive loss for the source language (which benefits from abundant data) and contrastive alignment for the target language (which operates under few-shot conditions). Extensive experiments on the X-stance, CIC, and VaxxStance datasets demonstrated that our method consistently outperforms competitive baselines including CCSD, CLKD, and JointCL, with the largest improvements reaching approximately 2\%--3\%. Furthermore, we empirically determined optimal configurations for In-Context Learning (8 examples) and support-query set construction (8 support samples per class with 4 query samples), balancing effectiveness and computational efficiency. Our qualitative analysis using Integrated Gradients illustrated that the model learns to focus on coherent phrasal reasoning rather than being misled by isolated keywords. While our framework still requires a small amount of labeled target-language data, it provides a practical and efficient solution for cross-lingual stance detection in realistic settings where deploying Large Language Models directly is infeasible due to high inference latency. Future work will explore zero-shot scenarios, alternative compact architectures (e.g., XLM-RoBERTa), multi-class stance detection, and dynamic adjustment of in-context examples based on topic complexity.

\vspace{6pt}

% \section*{Author Contributions}
% Conceptualization, Q.Z. ; Methodology, J.Y.; Validation, Q.Z., W.C., H.C. and S.T. ; Formal analysis, W.C. and H.C. ; Investigation, J.Y.; Resources, Q.Z.; Data curation, J.Y. ; Writing---original draft preparation, J.Y.; Writing---review and editing,  W.C., J.Y. and H.C.; Visualization, J.Y. and W.C.; Supervision, Q.Z. and W.C.; Project administration, Q.Z.; Funding acquisition, Q.Z. All authors have read and agreed to the published version of the manuscript.

% \section*{Data Availability}
% No new data were created or analyzed in this study. The publicly available datasets X-stance, CIC, and Vaxxstance, which support the findings of this study, are cited in the reference list.

% \section*{Conflicts of Interest}
% The authors declare no conflicts of interest.

% \begin{acks}
% This research was funded by the Anhui Province Philosophy and Social Science Planning Project (No. AHSKQ2024D044).
% \end{acks}

\bibliographystyle{unsrtnat}
\bibliography{ref}

@String{Computing = "Computing" }

@String{Computer = "{IEEE} Computer" }

@String{Springer = "Springer-Verlag" }

@inproceedings{recom1,
  title={Towards analyzing the bias of news recommender systems using sentiment and stance detection},
  author={Alam, Mehwish and Iana, Andreea and Grote, Alexander and Ludwig, Katharina and M{\"u}ller, Philipp and Paulheim, Heiko},
  booktitle={Companion proceedings of the web conference 2022},
  pages={448--457},
  year={2022}
}

@article{recom2,
  title={A survey on knowledge-aware news recommender systems},
  author={Iana, Andreea and Alam, Mehwish and Paulheim, Heiko},
  journal={Semantic Web},
  volume={15},
  number={1},
  pages={21--82},
  year={2024},
  publisher={SAGE Publications Sage UK: London, England}
}

@inproceedings{mono1,
  title={Cross-domain label-adaptive stance detection},
  author={Hardalov, Momchil and Arora, Arnav and Nakov, Preslav and Augenstein, Isabelle},
  booktitle={Proceedings of the 2021 conference on empirical methods in natural language processing},
  pages={9011--9028},
  year={2021}
}

@article{sdsurvey,
  title={Stance detection: A survey},
  author={K{\"u}{\c{c}}{\"u}k, Dilek and Can, Fazli},
  journal={ACM Computing Surveys (CSUR)},
  volume={53},
  number={1},
  pages={1--37},
  year={2020},
  publisher={ACM New York, NY, USA}
}

@article{socialmedia,
  title={Stance detection on social media: State of the art and trends},
  author={AlDayel, Abeer and Magdy, Walid},
  journal={Information Processing \& Management},
  volume={58},
  number={4},
  pages={102597},
  year={2021},
  publisher={Elsevier}
}

@inproceedings{mono3,
  title={Few-shot cross-lingual stance detection with sentiment-based pre-training},
  author={Hardalov, Momchil and Arora, Arnav and Nakov, Preslav and Augenstein, Isabelle},
  booktitle={Proceedings of the AAAI Conference on Artificial Intelligence},
  volume={36},
  number={10},
  pages={10729--10737},
  year={2022}
}

@inproceedings{zhang2023target,
  title={Target-oriented relation alignment for cross-lingual stance detection},
  author={Zhang, Ruike and Xu, Nan and Yang, Hanxuan and Tian, Yuan and Mao, Wenji},
  booktitle={Findings of the Association for Computational Linguistics: ACL 2023},
  pages={6391--6404},
  year={2023}
}

@inproceedings{zhang2024llm,
  title={An LLM-enabled knowledge elicitation and retrieval framework for zero-shot cross-lingual stance identification},
  author={Zhang, Ruike and Tian, Yuan and Wei, Penghui and Zeng, Daniel Dajun and Mao, Wenji},
  booktitle={Findings of the Association for Computational Linguistics: EMNLP 2024},
  pages={12253--12266},
  year={2024}
}

@inproceedings{zhang2023cross,
  title={Cross-lingual cross-target stance detection with dual knowledge distillation framework},
  author={Zhang, Ruike and Yang, Hanxuan and Mao, Wenji},
  booktitle={Proceedings of the 2023 conference on empirical methods in natural language processing},
  pages={10804--10819},
  year={2023}
}

@inproceedings{teachreason,
  title={Teaching small language models to reason},
  author={Magister, Lucie Charlotte and Mallinson, Jonathan and Adamek, Jakub and Malmi, Eric and Severyn, Aliaksei},
  booktitle={Proceedings of the 61st Annual Meeting of the Association for Computational Linguistics (Volume 2: Short Papers)},
  pages={1773--1781},
  year={2023}
}

@article{cotreason,
  title={Chain-of-thought prompting elicits reasoning in large language models},
  author={Wei, Jason and Wang, Xuezhi and Schuurmans, Dale and Bosma, Maarten and Xia, Fei and Chi, Ed and Le, Quoc V and Zhou, Denny and others},
  journal={Advances in neural information processing systems},
  volume={35},
  pages={24824--24837},
  year={2022}
}

@inproceedings{symbolic,
  title={Symbolic chain-of-thought distillation: Small models can also “think” step-by-step},
  author={Li, Liunian Harold and Hessel, Jack and Yu, Youngjae and Ren, Xiang and Chang, Kai-Wei and Choi, Yejin},
  booktitle={Proceedings of the 61st Annual Meeting of the Association for Computational Linguistics (Volume 1: Long Papers)},
  pages={2665--2679},
  year={2023}
}

@inproceedings{distillstepbystep,
  title={Distilling step-by-step! outperforming larger language models with less training data and smaller model sizes},
  author={Hsieh, Cheng-Yu and Li, Chun-Liang and Yeh, Chih-Kuan and Nakhost, Hootan and Fujii, Yasuhisa and Ratner, Alex and Krishna, Ranjay and Lee, Chen-Yu and Pfister, Tomas},
  booktitle={Findings of the Association for Computational Linguistics: ACL 2023},
  pages={8003--8017},
  year={2023}
}

@article{emergent,
  title={Emergent abilities of large language models},
  author={Wei, Jason and Tay, Yi and Bommasani, Rishi and Raffel, Colin and Zoph, Barret and Borgeaud, Sebastian and Yogatama, Dani and Bosma, Maarten and Zhou, Denny and Metzler, Donald and others},
  journal={arXiv preprint arXiv:2206.07682},
  year={2022}
}

@inproceedings{incontextsurvey,
  title={A survey on in-context learning},
  author={Dong, Qingxiu and Li, Lei and Dai, Damai and Zheng, Ce and Ma, Jingyuan and Li, Rui and Xia, Heming and Xu, Jingjing and Wu, Zhiyong and Chang, Baobao and others},
  booktitle={Proceedings of the 2024 conference on empirical methods in natural language processing},
  pages={1107--1128},
  year={2024}
}

@inproceedings{rethinkingincontext,
  title={Rethinking the role of demonstrations: What makes in-context learning work?},
  author={Min, Sewon and Lyu, Xinxi and Holtzman, Ari and Artetxe, Mikel and Lewis, Mike and Hajishirzi, Hannaneh and Zettlemoyer, Luke},
  booktitle={Proceedings of the 2022 conference on empirical methods in natural language processing},
  pages={11048--11064},
  year={2022}
}

@article{kdsurvey,
  title={Knowledge distillation: A survey},
  author={Gou, Jianping and Yu, Baosheng and Maybank, Stephen J and Tao, Dacheng},
  journal={International journal of computer vision},
  volume={129},
  number={6},
  pages={1789--1819},
  year={2021},
  publisher={Springer}
}

@article{hinton2015distilling,
  title={Distilling the knowledge in a neural network},
  author={Hinton, Geoffrey and Vinyals, Oriol and Dean, Jeff},
  journal={arXiv preprint arXiv:1503.02531},
  year={2015}
}

@article{meituan,
  title={From Reasoning LLMs to BERT: A Two-Stage Distillation Framework for Search Relevance},
  author={Xia, Runze and Ji, Yupeng and Zhou, Yuxi and Liu, Haodong and Zhang, Teng and Li, Piji},
  journal={arXiv preprint arXiv:2510.11056},
  year={2025}
}

@inproceedings{bert,
  title={Bert: Pre-training of deep bidirectional transformers for language understanding},
  author={Devlin, Jacob and Chang, Ming-Wei and Lee, Kenton and Toutanova, Kristina},
  booktitle={Proceedings of the 2019 conference of the North American chapter of the association for computational linguistics: human language technologies, volume 1 (long and short papers)},
  pages={4171--4186},
  year={2019}
}

@inproceedings{mbert,
  title={How multilingual is multilingual BERT?},
  author={Pires, Telmo and Schlinger, Eva and Garrette, Dan},
  booktitle={Proceedings of the 57th annual meeting of the association for computational linguistics},
  pages={4996--5001},
  year={2019}
}

@article{mismatch,
  title={Improving efficient neural ranking models with cross-architecture knowledge distillation},
  author={Hofst{\"a}tter, Sebastian and Althammer, Sophia and Schr{\"o}der, Michael and Sertkan, Mete and Hanbury, Allan},
  journal={arXiv preprint arXiv:2010.02666},
  year={2020}
}

@inproceedings{gao2021simcse,
  title={Simcse: Simple contrastive learning of sentence embeddings},
  author={Gao, Tianyu and Yao, Xingcheng and Chen, Danqi},
  booktitle={Proceedings of the 2021 conference on empirical methods in natural language processing},
  pages={6894--6910},
  year={2021}
}

@article{tian2019contrastive,
  title={Contrastive representation distillation},
  author={Tian, Yonglong and Krishnan, Dilip and Isola, Phillip},
  journal={arXiv preprint arXiv:1910.10699},
  year={2019}
}

@inproceedings{con3,
  title={A simple framework for contrastive learning of visual representations},
  author={Chen, Ting and Kornblith, Simon and Norouzi, Mohammad and Hinton, Geoffrey},
  booktitle={International conference on machine learning},
  pages={1597--1607},
  year={2020},
  organization={PmLR}
}

@inproceedings{single1,
  title={A dataset for multi-target stance detection},
  author={Sobhani, Parinaz and Inkpen, Diana and Zhu, Xiaodan},
  booktitle={Proceedings of the 15th conference of the European chapter of the association for computational linguistics: volume 2, short papers},
  pages={551--557},
  year={2017}
}

@inproceedings{single2,
  title={Semeval-2016 task 6: Detecting stance in tweets},
  author={Mohammad, Saif and Kiritchenko, Svetlana and Sobhani, Parinaz and Zhu, Xiaodan and Cherry, Colin},
  booktitle={Proceedings of the 10th international workshop on semantic evaluation (SemEval-2016)},
  pages={31--41},
  year={2016}
}

@inproceedings{single3,
  title={Stance detection in COVID-19 tweets},
  author={Glandt, Kyle and Khanal, Sarthak and Li, Yingjie and Caragea, Doina and Caragea, Cornelia},
  booktitle={Proceedings of the 59th Annual Meeting of the Association for Computational Linguistics and the 11th International Joint Conference on Natural Language Processing (Volume 1: Long Papers)},
  pages={1596--1611},
  year={2021}
}

@inproceedings{single4,
  title={Will-they-won’t-they: A very large dataset for stance detection on Twitter},
  author={Conforti, Costanza and Berndt, Jakob and Pilehvar, Mohammad Taher and Giannitsarou, Chryssi and Toxvaerd, Flavio and Collier, Nigel},
  booktitle={Proceedings of the 58th annual meeting of the association for computational linguistics},
  pages={1715--1724},
  year={2020}
}

@inproceedings{tan,
  title={Stance classification with target-specific neural attention networks},
  author={Du, Jiachen and Xu, Ruifeng and He, Yulan and Gui, Lin},
  booktitle={26th International Joint Conference on Artificial Intelligence, IJCAI 2017},
  pages={3988--3994},
  year={2017},
  organization={International Joint Conferences on Artificial Intelligence}
}

@inproceedings{before2,
  title={Connecting targets to tweets: Semantic attention-based model for target-specific stance detection},
  author={Zhou, Yiwei and Cristea, Alexandra I and Shi, Lei},
  booktitle={International Conference on Web Information Systems Engineering},
  pages={18--32},
  year={2017},
  organization={Springer}
}

@article{cross1,
  title={Zero-shot cross-lingual stance detection via adversarial language adaptation},
  author={Bharathi, A and Zubiaga, Arkaitz},
  journal={PeerJ Computer Science},
  volume={11},
  pages={e2955},
  year={2025},
  publisher={PeerJ Inc.}
}

@inproceedings{cross2,
  title={A Reinforcement Learning Framework for Cross-Lingual Stance Detection Using Chain-of-Thought Alignment},
  author={Li, Binghui and Zou, Minghui and Zhang, Xiaowang and Chen, Shizhan and Feng, Zhiyong},
  booktitle={Findings of the Association for Computational Linguistics: ACL 2025},
  pages={21674--21688},
  year={2025}
}

@article{kd1,
  title={BERTtoCNN: Similarity-preserving enhanced knowledge distillation for stance detection},
  author={Li, Yang and Sun, Yuqing and Zhu, Nana},
  journal={Plos one},
  volume={16},
  number={9},
  pages={e0257130},
  year={2021},
  publisher={Public Library of Science San Francisco, CA USA}
}

@article{llm1,
  title={Zero-shot stance detection in practice: insights on training, prompting, and decoding with a capable lightweight LLM},
  author={Aiyappa, Rachith and Senthilmani, Shruthi and An, Jisun and Kwak, Haewoon and Ahn, Yong-Yeol},
  journal={PeerJ Computer Science},
  volume={12},
  pages={e3540},
  year={2026},
  publisher={PeerJ Inc.}
}

@inproceedings{llm2,
  title={Stance reasoner: Zero-shot stance detection on social media with explicit reasoning},
  author={Taranukhin, Maksym and Shwartz, Vered and Milios, Evangelos},
  booktitle={Proceedings of the 2024 Joint International Conference on Computational Linguistics, Language Resources and Evaluation (LREC-COLING 2024)},
  pages={15257--15272},
  year={2024}
}

@inproceedings{llm3,
  title={Stance detection with collaborative role-infused llm-based agents},
  author={Lan, Xiaochong and Gao, Chen and Jin, Depeng and Li, Yong},
  booktitle={Proceedings of the international AAAI conference on web and social media},
  volume={18},
  pages={891--903},
  year={2024}
}

@inproceedings{vast,
  title={Zero-shot stance detection: A dataset and model using generalized topic representations},
  author={Allaway, Emily and McKeown, Kathleen},
  booktitle={Proceedings of the 2020 Conference on Empirical Methods in Natural Language Processing (EMNLP)},
  pages={8913--8931},
  year={2020}
}

@article{brown2020languageicl,
  title={Language models are few-shot learners},
  author={Brown, Tom and Mann, Benjamin and Ryder, Nick and Subbiah, Melanie and Kaplan, Jared D and Dhariwal, Prafulla and Neelakantan, Arvind and Shyam, Pranav and Sastry, Girish and Askell, Amanda and others},
  journal={Advances in neural information processing systems},
  volume={33},
  pages={1877--1901},
  year={2020}
}

@article{iclexplain,
  title={An explanation of in-context learning as implicit bayesian inference},
  author={Xie, Sang Michael and Raghunathan, Aditi and Liang, Percy and Ma, Tengyu},
  journal={arXiv preprint arXiv:2111.02080},
  year={2021}
}

@article{cotzero,
  title={Large language models are zero-shot reasoners},
  author={Kojima, Takeshi and Gu, Shixiang Shane and Reid, Machel and Matsuo, Yutaka and Iwasawa, Yusuke},
  journal={Advances in neural information processing systems},
  volume={35},
  pages={22199--22213},
  year={2022}
}

@article{xstance,
  title={X-stance: A multilingual multi-target dataset for stance detection},
  author={Vamvas, Jannis and Sennrich, Rico},
  journal={arXiv preprint arXiv:2003.08385},
  year={2020}
}

@inproceedings{cic,
  title={Multilingual stance detection in tweets: The Catalonia independence corpus},
  author={Zotova, Elena and Agerri, Rodrigo and Nu{\~n}ez, Manuel and Rigau, German},
  booktitle={Proceedings of the Twelfth Language Resources and Evaluation Conference},
  pages={1368--1375},
  year={2020}
}

@article{vaxxstance,
  title={VaxxStance: A dataset for cross-lingual stance detection on vaccines},
  author={Agerri, Rodrigo and Centeno, Roberto and Espinosa, Mar{\i}a and de Landa, Joseba Fernandez and Rodrigo, Alvaro},
  year={2021}
}

@inproceedings{crossnet,
  title={Cross-target stance classification with self-attention networks},
  author={Xu, Chang and Paris, C{\'e}cile and Nepal, Surya and Sparks, Ross},
  booktitle={Proceedings of the 56th Annual Meeting of the Association for Computational Linguistics (Volume 2: Short Papers)},
  pages={778--783},
  year={2018}
}

@inproceedings{jointcl,
  title={Jointcl: A joint contrastive learning framework for zero-shot stance detection},
  author={Liang, Bin and Zhu, Qinglin and Li, Xiang and Yang, Min and Gui, Lin and He, Yulan and Xu, Ruifeng},
  booktitle={Proceedings of the 60th annual meeting of the association for computational linguistics (volume 1: long papers)},
  pages={81--91},
  year={2022}
}

@inproceedings{clkd,
  title={Cross-lingual distillation for text classification},
  author={Xu, Ruochen and Yang, Yiming},
  booktitle={Proceedings of the 55th Annual Meeting of the Association for Computational Linguistics (Volume 1: Long Papers)},
  pages={1415--1425},
  year={2017}
}

@misc{qwen3.5,
    title  = {{Qwen3.5}: Towards Native Multimodal Agents},
    author = {{Qwen Team}},
    month  = {February},
    year   = {2026},
    url    = {https://qwen.ai/blog?id=qwen3.5}
}

@inproceedings{ye2024dual,
  title={Dual-path collaborative generation network for emotional video captioning},
  author={Ye, Cheng and Chen, Weidong and Li, Jingyu and Zhang, Lei and Mao, Zhendong},
  booktitle={Proceedings of the 32nd ACM International Conference on Multimedia},
  pages={496--505},
  year={2024}
}

@inproceedings{han2023text,
  title={Text style transfer with contrastive transfer pattern mining},
  author={Han, Jingxuan and Wang, Quan and Zhang, Licheng and Chen, Weidong and Song, Yan and Mao, Zhendong},
  booktitle={Proceedings of the 61st Annual Meeting of the Association for Computational Linguistics (Volume 1: Long Papers)},
  pages={7914--7927},
  year={2023}
}

@inproceedings{tian2023end,
  title={End-to-end aspect-based sentiment analysis with combinatory categorial grammar},
  author={Tian, Yuanhe and Chen, Weidong and Hu, Bo and Song, Yan and Xia, Fei},
  booktitle={Findings of the Association for Computational Linguistics: ACL 2023},
  pages={13597--13609},
  year={2023}
}

@article{fu2024sentiment,
  title={Sentiment-oriented transformer-based variational autoencoder network for live video commenting},
  author={Fu, Fengyi and Fang, Shancheng and Chen, Weidong and Mao, Zhendong},
  journal={ACM Transactions on Multimedia Computing, Communications and Applications},
  volume={20},
  number={4},
  pages={1--24},
  year={2024},
  publisher={ACM New York, NY}
}

@inproceedings{zhou2025hierarchical,
  title={Hierarchical Knowledge Distillation for Cross-Lingual Stance Detection},
  author={Zhou, Qiuli and Yao, Jingyuan and Tang, Shengeng and Chen, Weidong and Cheng, Lechao and Tang, Jun},
  booktitle={2025 4th International Conference on Artificial Intelligence, Human-Computer Interaction and Robotics (AIHCIR)},
  pages={1--5},
  year={2025},
  organization={IEEE}
}

@inproceedings{sundararajan2017axiomatic,
  title={Axiomatic attribution for deep networks},
  author={Sundararajan, Mukund and Taly, Ankur and Yan, Qiqi},
  booktitle={International conference on machine learning},
  pages={3319--3328},
  year={2017},
  organization={PMLR}
}

@article{TALLIP_1,
  title={Stance detection with a multi-target adversarial attention network},
  author={Sun, Qingying and Xi, Xuefeng and Sun, Jiajun and Wang, Zhongqing and Xu, Huiyan},
  journal={ACM Transactions on Asian and Low-Resource Language Information Processing},
  volume={22},
  number={2},
  pages={1--21},
  year={2022},
  publisher={ACM New York, NY}
}

@article{TALLIP_2,
  title={Knowledge-enhanced prompt-tuning for stance detection},
  author={Huang, Hu and Zhang, Bowen and Li, Yangyang and Zhang, Baoquan and Sun, Yuxi and Luo, Chuyao and Peng, Cheng},
  journal={ACM Transactions on Asian and Low-Resource Language Information Processing},
  volume={22},
  number={6},
  pages={1--20},
  year={2023},
  publisher={ACM New York, NY}
}

@article{TALLIP_3,
  title={Cross-lingual sentence embedding for low-resource Chinese-Vietnamese based on contrastive learning},
  author={Huang, Yuxin and Liang, Yin and Wu, Zhaoyuan and Zhu, Enchang and Yu, Zhengtao},
  journal={ACM Transactions on Asian and Low-Resource Language Information Processing},
  volume={22},
  number={6},
  pages={1--18},
  year={2023},
  publisher={ACM New York, NY}
}

@article{TALLIP_4,
  title={Target-oriented knowledge distillation with language-family-based grouping for multilingual nmt},
  author={Do, Heejin and Lee, Gary Geunbae},
  journal={ACM Transactions on Asian and Low-Resource Language Information Processing},
  volume={22},
  number={2},
  pages={1--18},
  year={2023},
  publisher={ACM New York, NY}
}

@article{TALLIP_5,
  title={More than syntaxes: Investigating semantics to zero-shot cross-lingual relation extraction and event argument role labelling},
  author={Wei, Kaiwen and Jin, Li and Zhang, Zequn and Guo, Zhi and Li, Xiaoyu and Liu, Qing and Feng, Weimiao},
  journal={ACM Transactions on Asian and Low-Resource Language Information Processing},
  volume={23},
  number={5},
  pages={1--21},
  year={2024},
  publisher={ACM New York, NY}
}

@article{TALLIP_6,
  title={Augmenting low-resource cross-lingual summarization with progression-grounded training and prompting},
  author={Ma, Jiushun and Huang, Yuxin and Wang, Linqin and Huang, Xiang and Peng, Hao and Yu, Zhengtao and Yu, Philip},
  journal={ACM Transactions on Asian and Low-Resource Language Information Processing},
  volume={23},
  number={9},
  pages={1--22},
  year={2024},
  publisher={ACM New York, NY}
}

@inproceedings{liu2024bootstrapping,
  title={Bootstrapping large language models for radiology report generation},
  author={Liu, Chang and Tian, Yuanhe and Chen, Weidong and Song, Yan and Zhang, Yongdong},
  booktitle={Proceedings of the AAAI Conference on Artificial Intelligence},
  volume={38},
  number={17},
  pages={18635--18643},
  year={2024}
}

@inproceedings{chen2022multi,
  title={Multi-attention network for compressed video referring object segmentation},
  author={Chen, Weidong and Hong, Dexiang and Qi, Yuankai and Han, Zhenjun and Wang, Shuhui and Qing, Laiyun and Huang, Qingming and Li, Guorong},
  booktitle={Proceedings of the 30th ACM International Conference on Multimedia},
  pages={4416--4425},
  year={2022}
}

@inproceedings{huang2025graph,
  title={Graph mixture of experts and memory-augmented routers for multivariate time series anomaly detection},
  author={Huang, Xiaoyu and Chen, Weidong and Hu, Bo and Mao, Zhendong},
  booktitle={Proceedings of the AAAI conference on artificial intelligence},
  volume={39},
  number={16},
  pages={17476--17484},
  year={2025}
}

@article{zhang2025creatidesign,
  title={Creatidesign: A unified multi-conditional diffusion transformer for creative graphic design},
  author={Zhang, Hui and Hong, Dexiang and Yang, Maoke and Cheng, Yutao and Zhang, Zhao and Chen, Weidong and Shao, Jie and Wu, Xinglong and Wu, Zuxuan and Jiang, Yu-Gang},
  journal={arXiv preprint arXiv:2505.19114},
  year={2025}
}

@article{li2024exploring,
  title={Exploring visual relationships via transformer-based graphs for enhanced image captioning},
  author={Li, Jingyu and Mao, Zhendong and Li, Hao and Chen, Weidong and Zhang, Yongdong},
  journal={ACM Transactions on Multimedia Computing, Communications and Applications},
  volume={20},
  number={5},
  pages={1--23},
  year={2024},
  publisher={ACM New York, NY}
}

@inproceedings{lin2024prompting,
  title={Prompting few-shot multi-hop question generation via comprehending type-aware semantics},
  author={Lin, Zefeng and Chen, Weidong and Song, Yan and Zhang, Yongdong},
  booktitle={Findings of the Association for Computational Linguistics: NAACL 2024},
  pages={3730--3740},
  year={2024}
}

@inproceedings{wang2023improving,
  title={Improving image captioning via predicting structured concepts},
  author={Wang, Ting and Chen, Weidong and Tian, Yuanhe and Song, Yan and Mao, Zhendong},
  booktitle={Proceedings of the 2023 conference on empirical methods in natural language processing},
  pages={360--370},
  year={2023}
}

@article{chen2023weakly,
  title={Weakly supervised text-based actor-action video segmentation by clip-level multi-instance learning},
  author={Chen, Weidong and Li, Guorong and Zhang, Xinfeng and Wang, Shuhui and Li, Liang and Huang, Qingming},
  journal={ACM Transactions on Multimedia Computing, Communications and Applications},
  volume={19},
  number={1},
  pages={1--22},
  year={2023},
  publisher={ACM New York, NY}
}

@inproceedings{chen2021cascade,
  title={Cascade cross-modal attention network for video actor and action segmentation from a sentence},
  author={Chen, Weidong and Li, Guorong and Zhang, Xinfeng and Yu, Hongyang and Wang, Shuhui and Huang, Qingming},
  booktitle={Proceedings of the 29th ACM International Conference on Multimedia},
  pages={4053--4062},
  year={2021}
}

@article{jin2024improving,
  title={Improving radiology report generation with multi-grained abnormality prediction},
  author={Jin, Yuda and Chen, Weidong and Tian, Yuanhe and Song, Yan and Yan, Chenggang},
  journal={Neurocomputing},
  volume={600},
  pages={128122},
  year={2024},
  publisher={Elsevier}
}

@article{ye2025improving,
  title={Improving video summarization by exploring the coherence between corresponding captions},
  author={Ye, Cheng and Chen, Weidong and Hu, Bo and Zhang, Lei and Zhang, Yongdong and Mao, Zhendong},
  journal={IEEE Transactions on Image Processing},
  year={2025},
  publisher={IEEE}
}

@article{song2025towards,
  title={Towards efficient partially relevant video retrieval with active moment discovering},
  author={Song, Peipei and Zhang, Long and Lan, Long and Chen, Weidong and Guo, Dan and Yang, Xun and Wang, Meng},
  journal={IEEE Transactions on Multimedia},
  year={2025},
  publisher={IEEE}
}

@inproceedings{li2025rethinking,
  title={Rethinking Pseudo Word Learning in Zero-Shot Composed Image Retrieval: From an Object-Aware Perspective},
  author={Li, Zhe and Zhang, Lei and Zhang, Kun and Chen, Weidong and Zhang, Yongdong and Mao, Zhendong},
  booktitle={Proceedings of the 48th International ACM SIGIR Conference on Research and Development in Information Retrieval},
  pages={833--843},
  year={2025}
}

@inproceedings{ye2025multi,
  title={Multi-round Mutual Emotion-Cause Pair Extraction for Emotion-Attributed Video Captioning},
  author={Ye, Cheng and Chen, Weidong and Song, Peipei and Liu, Xinyan and Zhang, Lei and Mao, Zhendong},
  booktitle={Proceedings of the 33rd ACM International Conference on Multimedia},
  pages={3320--3329},
  year={2025}
}

@inproceedings{wang2025combatting,
  title={Combatting Data Imbalance and Noise in Micro-Action Recognition},
  author={Wang, Chuang and Chen, Weidong and Cui, Xu and Zhao, Yiming and Qi, Zhaobo and Huang, Pengqi and Liu, Xinyan and Zhang, Weigang},
  booktitle={Proceedings of the 33rd ACM International Conference on Multimedia},
  pages={14229--14235},
  year={2025}
}

@inproceedings{wang2023contour,
  title={Contour-augmented concept prediction network for image captioning},
  author={Wang, Ting and Chen, Weidong and Li, Jingyu and Peng, Yixing and Mao, Zhendong},
  booktitle={International Conference on Artificial Neural Networks},
  pages={180--191},
  year={2023},
  organization={Springer}
}

@article{chen2026subjective,
  title={Subjective-objective Emotion Correlated Generation Network for Subjective Video Captioning},
  author={Chen, Weidong and Ye, Cheng and Song, Peipei and Zhang, Lei and Zhang, Yongdong and Mao, Zhendong},
  journal={IEEE Transactions on Image Processing},
  year={2026},
  publisher={IEEE}
}

@inproceedings{zhang2026stimuli,
  title={Stimuli-Aware Emotion Adaptor for Enhancing LLM in Affective Explanation Captioning},
  author={Zhang, Zhiyan and Song, Peipei and Hu, Jinpeng and Chen, Weidong and Ni, Lin and Yang, Xun},
  booktitle={ICASSP 2026-2026 IEEE International Conference on Acoustics, Speech and Signal Processing (ICASSP)},
  pages={10662--10666},
  year={2026},
  organization={IEEE}
}

@article{wang2026multi,
  title={A Multi-Agent Framework with Structured Reasoning and Reflective Refinement for Multimodal Empathetic Response Generation},
  author={Wang, Liping and Ye, Cheng and Chen, Weidong and Song, Peipei and Hu, Bo and Mao, Zhendong},
  journal={arXiv preprint arXiv:2604.18988},
  year={2026}
}

@article{chen2026creatiparser,
  title={CreatiParser: Generative Image Parsing of Raster Graphic Designs into Editable Layers},
  author={Chen, Weidong and Hong, Dexiang and Mao, Zhendong and Cheng, Yutao and Liu, Xinyan and Zhang, Lei and Zhang, Yongdong},
  journal={arXiv preprint arXiv:2604.19632},
  year={2026}
}

@article{chen2026face,
  title={FACE-net: Factual Calibration and Emotion Augmentation for Retrieval-enhanced Emotional Video Captioning},
  author={Chen, Weidong and Ye, Cheng and Mao, Zhendong and Song, Peipei and Liu, Xinyan and Zhang, Lei and Chang, Xiaojun and Zhang, Yongdong},
  journal={arXiv preprint arXiv:2603.17455},
  year={2026}
}

@inproceedings{qin2025query,
  title={Query-based Collaborative Multimodal Token Pruning for Audio-Visual Question Answering},
  author={Qin, Xilin and Hong, Dexiang and Chen, Weidong and Ye, Cheng and Liu, Xinyan and Song, Peipei and Zhang, Lei},
  booktitle={2025 4th International Conference on Artificial Intelligence, Human-Computer Interaction and Robotics (AIHCIR)},
  pages={1--6},
  year={2025},
  organization={IEEE}
}

@article{guo2025emoverse,
  title={EmoVerse: A MLLMs-Driven Emotion Representation Dataset for Interpretable Visual Emotion Analysis},
  author={Guo, Yijie and Hong, Dexiang and Chen, Weidong and She, Zihan and Ye, Cheng and Chang, Xiaojun and Mao, Zhendong},
  journal={arXiv preprint arXiv:2511.12554},
  year={2025}
}

@inproceedings{liu2025matching,
  title={Matching Street View and Satellite Images via Drone Imagery and Semantic Descriptions},
  author={Liu, Xinyan and Chen, Weidong and Qi, Zhaobo and Zhang, Beichen and Zhang, Weigang},
  booktitle={Proceedings of the 3rd International Workshop on UAVs in Multimedia: Capturing the World from a New Perspective},
  pages={4--9},
  year={2025}
}

@inproceedings{zhao2023difference,
  title={Difference-Aware Iterative Reasoning Network for Key Relation Detection},
  author={Zhao, Bowen and Chen, Weidong and Hu, Bo and Xie, Hongtao and Mao, Zhendong},
  booktitle={2023 IEEE International Conference on Multimedia and Expo (ICME)},
  pages={276--281},
  year={2023},
  organization={IEEE Computer Society}
}

\end{document}